%% file: main.tex
\pdfoutput=1
\documentclass[11pt]{article}

\usepackage[final]{acl}

\usepackage{times}
\usepackage{latexsym}
\usepackage{hyperref}
\usepackage[T1]{fontenc}
\usepackage[utf8]{inputenc}

\usepackage{microtype}

\usepackage{inconsolata}

\usepackage{graphicx}

\title{Instructions for *ACL Proceedings}

\usepackage{times}
\usepackage{latexsym}
\usepackage{enumitem}

\usepackage[T1]{fontenc}
\usepackage[utf8]{inputenc}

\usepackage{microtype}

\usepackage{inconsolata}

\usepackage{graphicx}

\usepackage{times}
\usepackage{latexsym}

\usepackage{booktabs}
\usepackage{siunitx}
\usepackage{amsmath,amssymb,amsthm}
\usepackage{multirow}
\usepackage{amsmath}
\usepackage{tikz}
\usepackage{array}
\newcolumntype{C}[1]{>{\centering\arraybackslash}p{#1}}
\usepackage{microtype}
\usepackage{subfig}
\usepackage{tabularx}

\title{The Geometry of Low-Resource Language Representations}
\author{Francois Meyer and Jan Buys \\
  Department of Computer Science \\
  University of Cape Town \\
  \texttt{francois.meyer@uct.ac.za, jbuys@cs.uct.ac.za}}

\begin{document}
\maketitle

\begin{abstract}
The performance gap between low- and high-resource languages in LLMs is widely known, but it remains unclear which internal model factors drive these disparities. In this paper, we characterise this gap through the lens of representational geometry. Comparing the geometric properties of hidden representations across 30 languages reveals that LLM geometry is systematically related to language data availability. The most consistent effect is in final layers, where low-resource languages exhibit representational degeneration. To counter this, we investigate the effectiveness of regularisation terms to penalise degeneration during continued pretraining (CPT). Experiments monolingually adapting 9 base LLMs to 10 African languages show that geometric regularisation successfully reduces representational degeneration during CPT. For larger models, cosine similarity-based regularisation marginally improves performance over vanilla CPT, with more consistent gains on the most challenging tasks. We establish that the representational geometry of low- and high-resource languages in LLMs is measurably distinct, and that targeted geometric intervention is a viable strategy for improving CPT for low-resource languages. 
\end{abstract}

\input{1intro}

\input{2background}

\input{3methodology}

\input{4analysis}

\input{5cpt}

\input{6conclusion}

\input{7limitations}

\appendix

\input{8appendix}

\end{document}

%% file: 1intro.tex
\section{Introduction}
\label{sec:introduction}

\begin{figure}[t]
 \vspace{-0.2cm}
  \includegraphics[trim={0.25cm 0.25cm 0 0cm},clip,width=\linewidth]{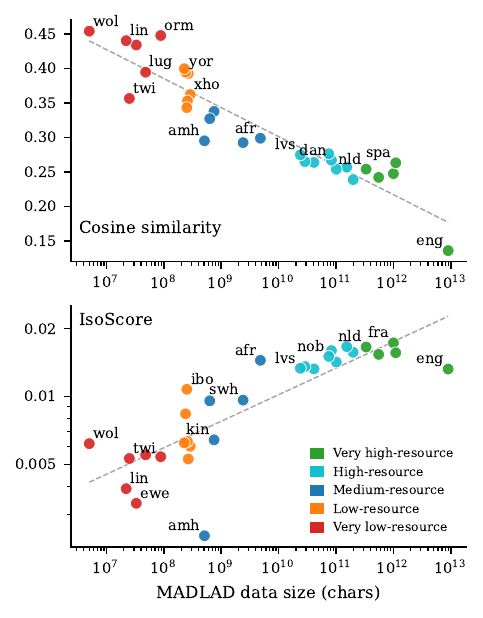}
  \vspace{-0.6cm}
  \caption{Geometric properties of final-layer token representations in Llama 3.1 8B (see Figures~\ref{fig:full_final_layer_cosine} and~\ref{fig:full_final_layer_isoscore} in the appendix for all models). Lower-resource languages exhibit greater representational degeneration. } 
  \vspace{-0.4cm}
  \label{fig:llama8b_final_layer_cosine}
\end{figure}

LLM capabilities have developed unevenly across languages, with performance in low-resource languages lagging far behind that of high-resource languages \citep{adelani-etal-2025-irokobench}.
This %disparity 
is primarily attributed to data scarcity \citep{joshi-etal-2020-state},
%and quality \citep{kreutzer-etal-2022-quality} 
with low-resource languages lacking sufficient pretraining data to enable generalisation across tasks. % and domains. 
From an interpretability perspective, it is unclear which internal model factors  
%-- the hidden representations and neural operations -- 
contribute to performance degradation for low-resource languages. 
% One way to study this is to analyse the differences between how LLMs process low- and high-resource languages, with the goal of identifying the underlying causes of cross-lingual performance gaps.
One way to study this is to compare how multilingual LLMs represent and process low- versus high-resource languages, with the goal of identifying the underlying causes of cross-lingual performance disparities.

In this work, we study the role of LLM geometry. We analyse how the geometric properties of language vector spaces (token embeddings across model layers) relate to data size. % and downstream performance. 
Our hypothesis is that vector spaces of low-resource languages are geometrically degenerate, that is, they fail to utilise the full representational capacity. 
% OLD
To test this, we use two metrics that characterise representational geometry from complementary perspectives: (1) \emph{Cosine similarity} measures how directionally distinct individual vectors are, with high similarities indicating representational collapse \citep{gao2018representation}. %, rajaee-pilehvar-2022-isotropy, godey2024why}. 
(2) \emph{Isotropy} measures how uniformly the dimensions are utilised -- anisotropic spaces have reduced effective dimensionality, which acts as a representational bottleneck \citep{rudman-etal-2022-isoscore}. %\citep{mu2018allbutthetop, rudman-etal-2022-isoscore, godey2024why}.
Cosine similarity is a local, relational metric, while isotropy measures global space utilisation. 
% OLD

% In this work, we investigate the role of representational geometry. We analyse how the geometric properties of language vector spaces (token embeddings across model layers) relate to data size. % and downstream performance. 
% Our hypothesis is that vector spaces of low-resource languages are geometrically degenerate, that is, they fail to utilise the full representational capacity. 
% To test this, we use two metrics that characterise geometry from complementary perspectives: cosine similarity measures how directionally distinct individual vectors are \citep{gao2018representation}, while isotropy measures how uniformly the dimensions of a vector space are utilised \citep{rudman-etal-2022-isoscore}. 
% % with high cosine similarity indicating representational collapse . 
% %  -- anisotropic spaces have reduced effective dimensionality, which acts as a representational bottleneck .
% % Cosine similarity is a local, relational metric, while isotropy measures global space utilisation \citep{rajaee-pilehvar-2021-cluster, rudman-etal-2022-isoscore}.
% Both can be used to quantify representational degeneration, but cosine similarity is a local, relational metric, while isotropy measures global space utilisation.

First, we conduct a multilingual analysis of LLM geometry. Using cosine similarity and isotropy, we compare how LLMs represent low- and high-resource languages. %modelacross different layers. 
%We measure cosine similarity and isotropy of embeddings for 18 languages with varying resource levels (English, French, and 16 African languages). 
Across 30 languages and 9 open models, 
%from Llama 3 \citep{llama3}, Gemma 3 \citep{gemmateam2025gemma3technicalreport}, and Qwen 3 \citep{yang2025qwen3technicalreport} series,
%(Llama 3 1/3/8B, Gemma 3 1/4/12B, and Qwen 3 1.7/4/8B), 
we observe a clear relationship between resource level and layer-wise geometry. The most consistent trend across metrics is in final layer representations: low-resource languages exhibit higher cosine similarity and anisotropy (as shown in Figure~\ref{fig:llama8b_final_layer_cosine}), both of which reflect representational degeneration. %We take this to be a partial explanation of why LLMs underperform on low-resource languages -- 
This shows that LLMs fail to learn geometrically rich representations for low-resource languages, limiting their ability to fully leverage their representational capacity.

% Next, we introduce consider how these insights can be used to improve performance for low-resource languages. 
% Assuming that the geometric properties observed in the final layer representations of high-resource languages are beneficial for performance, can we induce them for low-resource languages? 
% To do this, we employ two regularisation terms that augment the pretraining loss to explicitly shift final layer embeddings towards the properties observed for high-resource languages: a cosine similarity-based term that encourages local separability between token embeddings, and an isotropy-based term that encourages more uniform utilisation of the embedding space.

Next, we use these findings to augment continued pretraining (CPT), wherein a base LLM is adapted for a low-resource language. We employ two regularisation terms that augment the training objective to explicitly steer LLM representations toward desired geometric properties: CosReg \citep{gao2018representation}, which reduces pairwise cosine similarity to encourage more separable representations, and I-STAR \citep{rudman2024stable}, which increases isotropy to encourage more uniform use of the embedding space. 
These regularisation techniques have not previously been applied to CPT. 
%Based on our analysis, we configure both to penalise representational degeneration in final-layer representations. % arget the final layer and to encourage the geometric properties observed for high-resource languages.

% We test our geometric regularisation techniques in the context of continued pretraining (CPT), wherein a base LLM is adapted for a low-resource language through continued pretraining on a target-language corpus. 
We adapt 9 base LLMs through monolingual CPT for 10 African languages, comparing vanilla CPT to geometrically regularised CPT. Geometric regularisation effectively reduces representational degeneration, offering a simple method for targeted geometric intervention. Evaluating on IrokoBench \citep{adelani-etal-2025-irokobench}, CosReg marginally improves CPT for larger models (Llama 3.1 8B, Qwen 3 8B, and Gemma 3 4/12B), producing its largest gains on AfriMGSM, a highly challenging task for under-resourced languages. CosReg outperforms I-STAR, suggesting that local separability is a more influential factor in low-resource language modelling than overcoming global representational bottlenecks. % than global utilisation of the embedding space.. %, or at least more attainable with our methods.

%Mention actionable interprtability here?
In summary: we analyse how LLM geometry differs between low- and high-resource languages, observe clear differences in final-layer representations, and use geometrically regularised CPT to reduce these differences. Our findings highlight representational degeneration as a factor in the underperformance of low-resource languages, and present geometrically regularised CPT as a practical option for countering it.

%representational degeneration during CPT. 
%Our findings highlight the importance of LLM geometry in and 
%improve CPT for low-resource languages. 
%Our findings shed new light on the model-internal factors behind the underperformance of low-resource languages, and show that the geometric properties of internal LLM representations can be explicitly guided towards desirable characteristics. %Move to / incorporate in abstract?

%% file: 2background.tex
\section{Related Work}
\label{sec:related_work}
% What do I want to get across here?
% Representational degeneration is a thing and is measured in different ways (should I mention that there is some debate about metrics? - no, rather in section 3 when you introduce IsoScore)
% It has been linked to performance / softmax bottleneck, but there is some debate around performance.
% A few ways to addres representational collapse have been addressed.
% Some of these works have studied multilingual PLMs, although mostlt MLMs and sentence-level sementic tasks. The focus of this paper is on decoder-only LLMs and multilingual reasoning.

Previous studies on model geometry have repeatedly identified representational degeneration: embeddings fail to utilise the full capacity of the space. 
%Different metrics have been used to quantify this phenomenon. 
The most common metric for quantifying this phenomenon is cosine similarity, with high similarities reflecting embeddings clustered in a narrow region \citep{gao2018representation, godey2024why}.  
LM representation spaces have also been shown to be anisotropic: variance is dominated by a few axes and available dimensions are not utilised uniformly \citep{rajaee-pilehvar-2022-isotropy, rudman-etal-2022-isoscore}. 
%A related phenomenon is that of outlier dimensions \citep{kovaleva-etal-2021-bert, haemmerl-etal-2023-exploring}, high-variance features that models rely on disproportionately. % From a theoretical perspective, it is a proven
\citet{godey2024why} studied the connection between representational degeneration and model capacity, showing that final-layer representations of smaller LMs collapse onto a lower-dimensional subspace that acts as a representational bottleneck. %(the softmax bottleneck \citep{yang2018breaking}),

Several methods have been proposed to address representational collapse. A common strategy is to penalise degeneration during training with regularisation terms, based on cosine similarity \citep{gao2018representation}, isotropic metrics \citep{zhang-etal-2022-fine, ji-etal-2023-isotropic, rudman2024stable}, or contrastive learning \citep{zhang-etal-2022-fine, xiao-etal-2023-isotropy}. 
Findings have been mixed and there is no clear consensus about the relationship between geometric properties and downstream performance. Reducing representational degeneration improved performance in some contexts \citep{gao2018representation, rajaee-pilehvar-2022-isotropy, zhang-etal-2022-fine, haemmerl-etal-2023-exploring, ji-etal-2023-isotropic}, but in others it had no effect or even degraded it \citep{rajaee-pilehvar-2021-fine-tuning, kovaleva-etal-2021-bert, zhang-etal-2022-fine, ding-etal-2022-isotropy, rudman2024stable}.

Some of the works outlined above have studied multilingual models \citep{rajaee-pilehvar-2022-isotropy, haemmerl-etal-2023-exploring, ji-etal-2023-isotropic} and designed methods to counter representational collapse in multilingual settings, such as training on parallel corpora \citep{haemmerl-etal-2023-exploring} or code-switched data \citep{ji-etal-2023-isotropic}. However, these studies focussed on masked LMs and sentence-level semantic tasks. 
%More broadly, while cross-lingual geometry has been widely studied by measuring alignment \emph{between} language representations \citep{hammerl-etal-2024-understanding}, less attention has been paid to 
%characterising the geometry of individual language representation spaces and comparing their properties across languages.
%comparing the geometric properties \emph{within} each language's representation space. 
In modern decoder-only LLMs, the relationship between data scarcity and representational geometry has not been studied. The question of whether low-resource languages have systematically different representational geometry from high-resource ones, and whether this can be addressed through geometric regularisation, remains unexplored. In this work we investigate both.

%% file: 3methodology.tex
\section{Methodology}
\label{sec:methodology}

This paper presents two complementary studies. 
First, we systematically compare the geometry of LLM representation spaces across low- and high-resource languages (Section~\ref{sec:analysis}).
% Second, based on these findings, we propose geometric regularisation techniques to improve CPT for low-resource languages (Section~\ref{sec:cpt_results}). 
Second, based on these findings, we test regularisation terms that counteract the geometric deficiencies observed for low-resource languages (Section~\ref{sec:cpt_results}).
Our overall approach follows the paradigm of actionable interpretability \citep{mosbach-etal-2024-insights, orgad2026actionable}: using insights from interpretability analysis to inform  modelling decisions.\footnote{Code for our geometric analysis and regularised CPT is available at \href{https://github.com/francois-meyer/cpt-geometry}{\tt github.com/francois-meyer/cpt-geometry}.} In this section, we describe the experimental setup of our geometric analysis (Section~\ref{subsec:geometric_analysis}) and our CPT experiments with geometric regularisation (Sections~\ref{subsec:cpt}--\ref{subsec:evaluation}).

\subsection{Geometric Analysis}
\label{subsec:geometric_analysis}

Our starting point is to compare the geometric properties of representation spaces across different languages, models, and layers. We study 30 languages, selected to span a wide range of data availability. To quantify per-language data scarcity, we use corpus size in the MADLAD-400 dataset \citep{kudugunta2023madlad} as a proxy. MADLAD is a popular multilingual pretraining corpus based on Common\-Crawl. Although it does not reflect the exact training mixtures of the models we analyse (their pretraining corpora are not publicly disclosed), it provides a reasonable approximation of relative web-scale data availability across languages. 
Based on MADLAD data size, we group languages into five tiers of ``resourcedness'':
\begin{itemize}[leftmargin=*, noitemsep]
    \item \textbf{Very high} ({>}200B chars): English, Spanish, French, Italian, Dutch
    \item \textbf{High} ({>}5B chars): Estonian, Turkish, Swedish, Finnish, Danish, Norwegian, Lithuanian, Latvian
    \item \textbf{Medium} (500M--5B chars): Afrikaans, Kiswahili, Kinyarwanda, Hausa, Amharic
    \item \textbf{Low} (100M--500M chars): isiXhosa, chi\-Shona, isiZulu, Igbo, Yor\`ub\'a, Sesotho
    \item \textbf{Very low} ({<}100M chars): Oromo, Luganda, Ewe, Twi, Lingala, Wolof
\end{itemize}
% (English, French, Kiswahili, Kinyarwanda, Hausa, Amharic, isiXhosa, chiShona, isiZulu, Igbo, Yorùbá, Sesotho, Oromo, Luganda, Ewe, Twi, Lingala, Wolof)
%We compare the representation spaces of these languages across nine open models: 
% \begin{itemize}[noitemsep]
%     \item Llama 3.2 (1B, 3B), 3.1 (8B) \citep{llama3techreport}
%     \item Gemma 3 (1B, 4B, 12B) \citep{gemma3techreport}
%     \item Qwen 3 (1.7B, 4B, 8B) \citep{qwen3techreport} 
% \end{itemize}
We compare the representation spaces of these languages across 9 open models: Llama 3.2 (1B, 3B), 3.1 (8B) \citep{llama3techreport}, Gemma 3 (1B, 4B, 12B) \citep{gemma3techreport}, and Qwen 3 (1.7B, 4B, 8B) \citep{qwen3techreport}.
% Previous work \citep{yu2026afriquellmdatamixingmodel} has shown that the Gemma 3 series \citep{gemmateam2025gemma3technicalreport} has strong multilingual capabilities, followed by Qwen 3 \citep{yang2025qwen3technicalreport} and Llama 3 \citep{llama3}. 
Our setup spans languages with varying levels of data availability, models with varying levels of multilingual competence, and models ranging from 1B to 12B parameters.

We use two metrics to quantify vector space geometry: cosine similarity and isotropy. 
For each language, we pass sentences from FLORES \citep{goyal-etal-2022-flores} through a model and compute both metrics based on token embeddings from each layer.

\subsubsection{Cosine similarity} 
\label{subsec:cosine_similarity}
For layer $l$, we compute the average pairwise cosine similarity as           
{
\setlength{\abovedisplayskip}{5pt}
\setlength{\belowdisplayskip}{6pt}
\begin{align}
    \mathrm{CosSim}(l) = \frac{1}{N} \sum_{(i,j)} \frac{\mathbf{h}_i \cdot  \mathbf{h}_j}{|\mathbf{h}_i| |\mathbf{h}_j|},
\end{align}
}where $i, j$ correspond to randomly sampled pairs of tokens from FLORES sentences, and $\mathbf{h}_i, \mathbf{h}_j$ are hidden representations of these tokens in layer $l$. We sample $N=1000$ token pairs per layer. 

Cosine similarity is a local, relational metric, quantifying how directionally distinct vectors are. Averaged across a sample of embeddings, it estimates how geometrically concentrated or dispersed the representations of a language are. 
For LLM representations, cosine similarity typically ranges from 0 to 1. Low values indicate well-separated embeddings, while high values indicate embeddings collapsed onto a narrow, high-dimensional cone.  %Cosine similarity is of interest in studying LLM geometry LLMs because dot products (proportional to cosine similarity under normalisation) are central to Transformer computations. 

\subsubsection{Isotropy} 
\label{subsec:isotropy}
Average pairwise cosine similarity is often used as a measure of isotropy. However, \citet{rudman-etal-2022-isoscore} argue that it fails to capture the true definition of isotropy: how uniformly variance is distributed across the dimensions of a vector space. 

They propose IsoScore, a metric that quantifies the degree to which the variance of a set of embeddings is uniformly distributed across all dimensions. Based on eigenvalues of the covariance matrix, it quantifies how evenly variance is spread across dimensions. %We refer the reader to \citet{rudman-etal-2022-isoscore} for details).
Unlike cosine similarity, it is invariant to transformations that do not change isotropy (e.g. mean-based shifting). IsoScore measures global space utilisation, while cosine similarity captures local separability between individual vectors.

Following  \citet{rudman-etal-2022-isoscore}, we compute IsoScore for layer $l$, over a point cloud $X$ of $N=5000$ randomly sampled token embeddings from FLORES sentences, as
{
\setlength{\abovedisplayskip}{5pt}
\setlength{\belowdisplayskip}{6pt}
\begin{align}
    \scalebox{0.90}{$\displaystyle
        \mathrm{IsoScore}(l) = \frac{\left(d - \delta(X)^2 (d - \sqrt{d})\right)^2 - d}{d(d-1)},
    $}
\end{align}
}
where $d$ is the embedding dimension and $\delta(X) \in [0,1]$ is the isotropy defect, measuring how far the variance distribution of $X$ deviates from uniformity (we refer the reader to \citet{rudman-etal-2022-isoscore} for a more detailed presentation of the IsoScore metric). A score of 1 indicates perfectly isotropic representations (all dimensions utilised evenly), while lower values reflect increasing anisotropy.

\subsection{CPT with Geometric Regularisation}
\label{subsec:cpt}

Our geometric analysis, outlined above, is an interpretability study that aims to highlight differences in LLM geometry across languages. Next, we consider a practical application where these differences might provide actionable insights: continued pretraining (CPT). In CPT, a pretrained checkpoint is further trained on a target language corpus under the model's original pretraining objective. It is used to adapt pretrained models to a specific language or limited set of languages. 

%By initialising from a pretrained checkpoint, CPT can leverage cross-lingual knowledge from the original pretraining phase adapt its parameters to model target languages. For low-resource languages, this often improves over training from scratch, both in terms of sample-efficiency and downstream performance. 
%As a result, 
CPT is the standard approach for developing high-quality LMs for low-resource languages, particularly in settings where training from scratch is computationally infeasible. 
For African languages, CPT has produced state-of-the-art results across different architectures, from masked language models \citep{alabi-etal-2022-adapting}, to encoder–decoder models \citep{oladipo-etal-2023-better}, and more recently decoder-based LLMs \citep{buzaaba2025lughallamaadaptinglargelanguage, yu2026afriquellmdatamixingmodel}. 
% However, %despite being widely adopted, 
% CPT remains under-studied as a training phase. 
% The mechanisms underlying language adaptation are unknown -- which model-internal changes occur over the course of CPT to improve performance for a target language?
% In this paper, we consider this question from the perspective of LLM geometry by studying how the geometric properties of target-language representations change during CPT. 
% Despite this success, there has been little work on improving the CPT procedure itself — most approaches simply minimise the standard language modelling loss on the target-language corpus without modification. 

  \begin{figure*}[t!]
% \vspace{-0.5cm}
  \includegraphics[trim={0cm 0 0 0cm},clip,width=\linewidth]{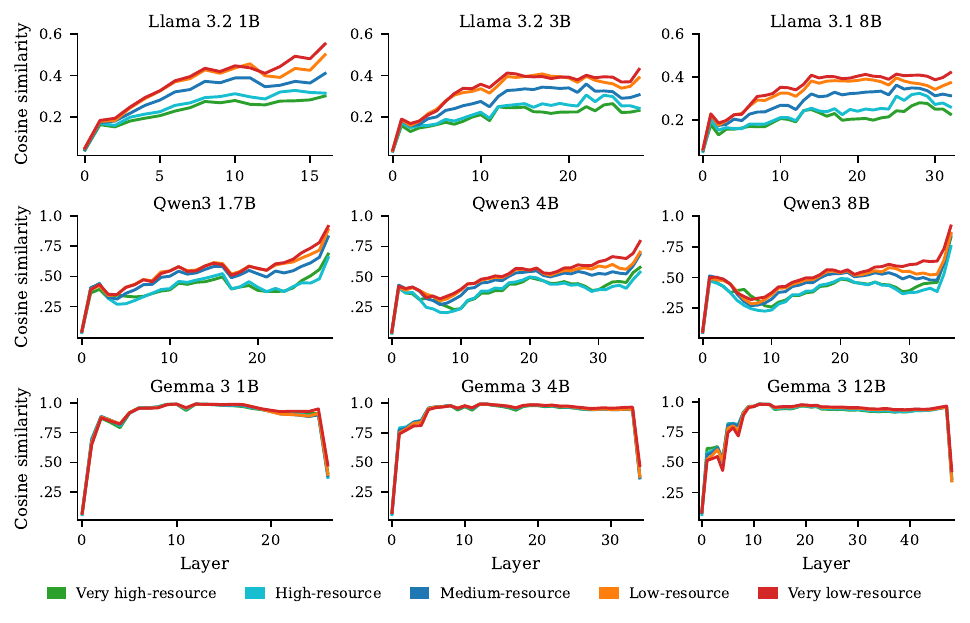}
  \vspace{-0.75cm}
  \caption{Pairwise cosine similarity of representations across model layers, calculated as specified in Section~\ref{subsec:cosine_similarity} and averaged across languages within each tier of data availability.} 
  \vspace{-0.2cm}
  \label{fig:cosine_layers}
\end{figure*}

We apply monolingual CPT to all 9 LLMs from our geometric analysis, on corpora from 10 African languages: Kinyarwanda, Hausa, Amharic, isi\-Xhosa, chiShona, isiZulu, Igbo, Yorùbá, Sesotho, and Oromo. For CPT we use the WURA corpus \citep{oladipo-etal-2023-better}, a high-quality filtered subset of mC4. We select languages in WURA that are in our very low-, low-, and medium-resource tiers, as this is where CPT offers most benefit (see Table~\ref{tab:wura_sizes} in the Appendix for corpus statistics). 
%pretraining corpus \citep{oladipo-etal-2023-better}, which we use for CPT. % (see Table~\ref{tab:wura_sizes} in the appendix for corpus sizes).
% and evaluation ? IrokoBench \citep{adelani-etal-2025-irokobench}
%All these languages are included  
For each model and language, we continue pretraining the base LLM\footnote{We use base models, instead of instruction-tuned versions, in all our experiments. CPT extends pretraining, so it is more naturally applied to base models that have not undergone additional post-training.} 
%instruction tuning may change representational geometry in ways that confound our analysis.
for 1 epoch on the target-language corpus. %(hyperparameters included in the appendix)
Our baseline models are adapted with vanilla CPT, in which the standard LM loss is minimised over the target-language corpus. Our hyperparameter tuning process and final CPT configuration is detailed in Appendix~\ref{appendix:cpt_hyperparams}.

% \subsection{Geometric Regularisation}
% \label{subsec:geometric_reg}

Vanilla CPT is a strong baseline for low-resource language adaptation. 
% Our goal is to improve on it by explicitly shaping the geometry of learned representations during CPT. 
Because CPT updates model parameters on target-language data, it provides an opportunity to correct the geometric deficiencies inherited from pretraining. We explicitly encourage this by augmenting the CPT objective with regularisation terms that penalise representational degeneration.
%To do this, we augment the CPT objective with regularisation terms that penalise representational degeneration. 
We test two regularisation strategies, each targeting one of the geometric metrics from our analysis. We tune hyperparameters for both regularisers, as detailed in Appendix~\ref{appendix:reg_hyperparams}.

\paragraph{CosReg} Following \citet{gao2018representation}, we augment the training objective with a penalty based on the mean pairwise cosine similarity of all $M$ token representations at layer $l$ in a batch:                                                                                                                                     
  \begin{align}                                                            
      \mathcal{L}_{\text{CosReg}} = \mathcal{L}_{\text{NLL}} + \lambda \cdot \frac{1}{M^2} \sum_{i} \sum_{j \neq i} \frac{\mathbf{h}_i \cdot \mathbf{h}_j}{|\mathbf{h}_i|             
  |\mathbf{h}_j|},                                                                                                           \label{eq:cos_reg}                                  
  \end{align}                                                                                                                                                                       
 where ${\mathbf{h}_1, \ldots, \mathbf{h}_M}$ are token representations at layer $l$ in a training mini-batch. 
We set $\lambda = 1$ to penalise high pairwise cosine similarity and encourage more separable representations.
% The sign of $\lambda$ determines the direction of the intervention: 
%   positive values encourage lower cosine similarity (more separated representations), while negative values encourage higher cosine similarity. The magnitude of $\lambda$ controls 
%   the strength of the regularisation relative to the language modelling objective. The regularisation can be applied to a single layer or multiple layers simultaneously.

\paragraph{I-STAR} \citet{rudman2024stable} propose I-STAR, a regularisation term that estimates isotropy from the representations of a batch. 
It is based on IsoScore$^\star$, a differentiable variant of IsoScore that can be optimised during training. They apply it during masked LM finetuning, but suggest its use in pretraining as a promising direction for future work. We adopt I-STAR for CPT of decoder LLMs. 

At each training step, we compute IsoScore$^\star$ for a randomly sampled batch subset of $M'=1{,}000$ token embeddings at layer $l$, and augment the training objective as follows:
\begin{align}
    \mathcal{L}_{\text{I-STAR}} = \mathcal{L}_{\text{NLL}} + \lambda \cdot (1 - \text{IsoScore}^\star).
    \label{eq:istar_reg} 
\end{align}
We set $\lambda = 1$ to penalise anisotropy and encourage more even utilisation of dimensions. \citet{rudman2024stable} found that \emph{encouraging} anisotropy with I-STAR ($\lambda < 0$) produced better results in the context of masked LM finetuning. We did test negative values of $\lambda$ during hyperparameter tuning (Appendix~\ref{appendix:reg_hyperparams}) but found no benefit for CPT. 

% The sign of $\lambda$ determines the direction of the intervention: positive values encourage greater isotropy, while negative values encourage greater anisotropy. The magnitude of $\lambda$ controls the strength of the regularisation relative to the LM loss. As with CosReg, the regularisation can be applied to a single layer or multiple layers simultaneously. However, the cost of computing IsoScore$^\star$ makes it computationally challenging to scale multi-layer I-STAR regularisation to CPT.

\vspace{0.5em}
% In both cases, the specific configuration -- which layer(s) to regularise, and the sign and magnitude of $\lambda$ -- will be determined by the findings of our geometric analysis (Section~\ref{sec:analysis}). We will select configurations that encourage, for low-resource languages, the geometric properties observed in high-resource language representations.

\subsection{Evaluation}
\label{subsec:evaluation}

To test the impact of geometric intervention, we compare target-language performance after vanilla CPT and geometrically regularised CPT. We evaluate adapted models on target-language data sets in IrokoBench \citep{adelani-etal-2025-irokobench}, a human-translated benchmark for African languages that includes natural language inference (AfriXNLI), multi-choice knowledge-based QA (AfriMMLU), and mathematical reasoning (AfriMGSM). We evaluate with LM Evaluation Harness \citep{eval-harness} in zero-shot and few-shot settings (5-shot for AfriMMLU and AfriXNLI, 8-shot for AfriMGSM), using prompt templates from \citet{adelani-etal-2025-irokobench}.

%% file: 4analysis.tex
\section{Multilingual Geometric Analysis}
\label{sec:analysis}

% What do I want to get across here?
% 1. There are clear, consistent patterns and differences between high and low-resource languages. Level of resource ~ geometric properties.
% 2. Final layer behaviour is especially notable different -- low-resource languages exhibit representational degeneration -- AND is particularly correlated with level of resourcedness.
% 3. Other factors -- gemma looks very different to other models (why?), Amharic is an outlier (why?), isotropy is so close (why?)

We present our geometric analysis (Section~\ref{subsec:geometric_analysis}) of nine models across 30 languages.

\subsection{Geometry Across Languages and Layers}
\label{subsec:geometry_across_languages_and_layers}
We observe clear differences between the geometric properties of high- and low-resource languages. Figure~\ref{fig:cosine_layers} plots average cosine similarity, grouped by data tier. Lower-resource languages have higher cosine similarity across most layers, while higher-resource language representations are more dispersed, indicating better local separability. 
% LLMs fail to learn sufficiently discriminative representations for low-resource languages, which we take as a partial explanation for their weaker performance. Advanced LLM capabilities require rich, well-separated representations, and data scarcity prevents LLMs from developing them. 

This trend is not only a binary distinction between low- and high-resource languages. In many layers, language data size and cosine similarity are strongly correlated: the lower-resource a language is, the higher its cosine similarity.  
This relationship can be seen in the Llama and Qwen plots of Figure~\ref{fig:cosine_layers}, where cosine similarity decreases monotonically with data tier for many layers (from very low-resource to very high-resource). To quantify this relationship, we compute Spearman $\rho$ correlation coefficients between MADLAD data size and cosine similarity for each layer. Table~\ref{tab:spearman_combined} reports $\rho$ for evenly-spaced layers across all models. Deeper layers have strong negative correlations (similarity decreases as language ``resourcedness'' increases). 

We also observe clear differences between the isotropy of high- and low-resource languages in some layers. Low-resource language representation spaces tend to be more anisotropic. This is hard to visualise because IsoScore differences are small (see Figure~\ref{fig:isoscore_layers} in the appendix), but the correlation analysis between data size and IsoScore in Table~\ref{tab:spearman_combined} shows a clear pattern.  Input layers have weak positive correlations, middle-layers have inconsistent correlations, and final layers have the strongest, positive correlations  (IsoScore increases as language ``resourcedness'' increases).
%\footnote{The variation in layer-wise geometry across model families, notable in Figure~\ref{fig:cosine_layers}, warrants further study (see Section~\ref{sec:limitations}).} 
%A high $\rho$ value means low-resource languages tend to have to have more isotropic representations.
% In final layers, higher-resource languages tend to have to have more isotropic representations. 
% Across all languages, a similar layer-wise trend emerges: 

One language emerged as an outlier: Amharic consistently deviated from what its data availability would predict (as shown in Figure~\ref{fig:llama8b_final_layer_cosine}). The layer-wise geometric profiles of Amharic were also markedly different from other languages in its resource tier. We attribute this to Amharic's use of the Ge'ez script, which sets it apart from the remaining languages, all of which use Latin script.  
Amharic shares no subword tokens with our other languages. 
Its representations originate from a geometrically isolated set of input embeddings and remain geometrically distinct throughout model layers. This highlights that resource level is not the only factor influencing LLM geometry and that linguistic typology can also play a role. 

\input{tables/spearman_combined}

\subsection{Final Layer Degeneration}
% Conclude with a paragraph motivating next section...

The most consistent trend across metrics and models concerns final layer representations, where low-resource languages reliably exhibit greater representational degeneration than high-resource ones. As shown in Figure~\ref{fig:llama8b_final_layer_cosine} for Llama 3.1 8B (and Figures~\ref{fig:full_final_layer_cosine} and~\ref{fig:full_final_layer_isoscore} in the appendix for all models), final-layer degeneration worsens as language data decreases. 
While absolute measures of degeneration actually improve for some models/languages in final layers (e.g. cosine similarity of Gemma models in Figure~\ref{fig:cosine_layers}, IsoScore of most models in Figure~\ref{fig:isoscore_layers}), cross-lingual geometric differences persist and, in many  instances, become more pronounced. 
The last column of Table~\ref{tab:spearman_combined} shows that the final layer is the only layer where representational degeneration correlates with data scarcity across all models and both metrics ($\rho$ is consistently negative for cosine similarity and positive for IsoScore).
%  , the relationship between data size and representational degeneration is strongest in the final layer. 
% Seeing the final layer is where data scarcity has the greatest influence on LLM geometry, in the next section we focus on addressing degeneration in final-layer representations.

%so this is where data size has the greatest influence on LLM isotropy. 
% As shown , the degree of degeneration is inversely proportional to language data size: the less training data available for a language, the more final-layer degeneration (higher cosine similarity and lower IsoScore). 

% Final-layer representations are also distinct in other respects. 

% As shown in Figure~\ref{fig:cosine_layers}, final-layer geometry is an exception to some of the trends observed across the rest of the model. For some models () final-layer representations 

% Most notably, final-layer cosine similarity of high-resource languages \emph{decreases} from preceding layers, in contrast to that of low-resource languages, which see  
% \citet{godey2024why} reported a  

\subsection{Geometry Across Models}
\label{subsec:geometry_across_models}

Figures~\ref{fig:cosine_layers} and~\ref{fig:isoscore_layers} also show geometric differences between model families. 
\citet{yu2026afriquellmdatamixingmodel} showed that Gemma 3 has strong multilingual capabilities, followed by Qwen 3 and Llama 3.
This is reflected in LLM geometry -- differences between lower and higher-resource languages shrink as multilingual coverage expands from Llama, to Qwen, and Gemma. Table~\ref{tab:spearman_combined} also shows that the correlation between data size and final-layer degeneration is strongest in Llama, weaker in Qwen, and weakest in Gemma (this holds for both cosine similarity and IsoScore). However, broader multilingual coverage does not eliminate cross-lingual disparities completely. Even in Gemma, where geometric differences are hard to discern in our plots, data scarcity still consistently correlates with representational degeneration in deeper layers.

\input{tables/geometry_change_final_layer}

\input{tables/performance_summary}

A notable feature of our plots is the distinct geometric profile of Gemma compared to Llama and Qwen. In Figure~\ref{fig:cosine_layers}, Gemma has high cosine similarity across most layers and a dramatic drop in the final layer. 
%Determining the cause of this is beyond the scope of our work, but there are several architectural choices unique to Gemma 3 that could plausibly contribute. 
This likely reflects differences in architecture and pretraining between the model families. 
For example, unlike Llama and Qwen, Gemma applies layer normalisation both before and after each sublayer, which may constrain angular variation across layers. %prevent representations from becoming more separable in intermediate layers.
Gemma also uses a larger vocabulary, %(262K tokens, compared to 128K for Llama 3 and 152K for Qwen 3), 
which requires more discriminative final-layer representations to predict next tokens.
Other factors in Gemma pretraining, such as knowledge distillation and multi-modal pretraining, might also influence model geometry. 
Given architectural differences and varying pretraining choices, it is unsurprising that models have distinct geometric profiles. 
A detailed investigation of these effects is outside the scope of this work. 
Instead, we focus on what is consistent across all models: low-resource languages exhibit greater representational degeneration, regardless of architecture.

%% file: tables/spearman_combined.tex
\begin{table}[t]
\centering
\small
\begin{tabular}{lrrrrr}
\toprule
Model & 0\% & 25\% & 50\% & 75\% & 100\% \\
\midrule
\multicolumn{6}{l}{\textit{Cosine similarity}} \\
\midrule
Llama 3.2 1B &  \textbf{-0.52} & \textbf{-0.84} & \textbf{-0.89} & \textbf{-0.88} & \textbf{-0.88} \\
Llama 3.2 3B & -0.24 & \textbf{-0.87} & \textbf{-0.89} & \textbf{-0.85} & \textbf{-0.90} \\
Llama 3.1 8B &  \textbf{-0.52} & \textbf{-0.93} & \textbf{-0.93} & \textbf{-0.90} & \textbf{-0.95} \\
Qwen3 1.7B &  \textbf{-0.56} & \textbf{-0.76} & \textbf{-0.85} & \textbf{-0.82} & \textbf{-0.73} \\
Qwen3 4B &  \textbf{-0.56} & \textbf{-0.78} & \textbf{-0.91} & \textbf{-0.85} & \textbf{-0.74} \\
Qwen3 8B &  \textbf{-0.56} & \textbf{-0.73} & \textbf{-0.89} & \textbf{-0.89} & \textbf{-0.58} \\
Gemma 3 1B & -0.27 & 0.58 & 0.47 & -0.17 & -0.42 \\
Gemma 3 4B &  \textbf{-0.54} & 0.80 & \textbf{-0.75} & \textbf{-0.83} & -0.40 \\
Gemma 3 12B &  \textbf{-0.52} & 0.38 & \textbf{-0.71} & \textbf{-0.81} & -0.25 \\
\midrule
\multicolumn{6}{l}{\textit{IsoScore}} \\
\midrule
Llama 3.2 1B & 0.20 & 0.04 & 0.20 & 0.29 & \textbf{0.85} \\
Llama 3.2 3B & 0.15 & 0.10 & 0.13 & 0.36 & \textbf{0.88} \\
Llama 3.1 8B & \textbf{0.72} & -0.26 & 0.06 & 0.07 & \textbf{0.85} \\
Qwen3 1.7B & 0.39 & 0.09 & 0.26 & 0.30 & 0.48 \\
Qwen3 4B & 0.47 & -0.10 & -0.11 & -0.19 & \textbf{0.73} \\
Qwen3 8B & 0.45 & -0.33 & -0.17 & -0.39 & \textbf{0.70} \\
Gemma 3 1B & -0.04 & -0.06 & \textbf{0.83} & \textbf{0.76} & \textbf{0.53} \\
Gemma 3 4B & 0.03 & -0.29 & \textbf{0.53} & \textbf{0.68} & 0.47 \\
Gemma 3 12B & 0.30 & -0.27 & \textbf{0.55} & \textbf{0.71} & 0.25 \\
\bottomrule
\end{tabular}
\caption{Spearman $\rho$ between MADLAD size and geometric metrics across layers (0\% = input embeddings, 100\% = final layer, remaining columns are evenly-spaced hidden layers). We \textbf{boldface} $\rho < -0.5$ for cosine similarity and $\rho > 0.5$ for IsoScore.}
\label{tab:spearman_combined}
\end{table}

%% file: tables/geometry_change_final_layer.tex
\begin{table}[t]
\setlength{\tabcolsep}{3pt}
\centering
\small
\begin{tabular}{l rrr rrr}
\toprule
& \multicolumn{3}{c}{$\Delta$ Cosine similarity} & \multicolumn{3}{c}{$\Delta$ IsoScore} \\
\cmidrule(lr){2-4} \cmidrule(lr){5-7}
Model & CPT & +CR & +IS & CPT & +CR & +IS \\
\midrule
Llama 3.2 1B & -.096 & -.278 & -.089 & .000 & -.004 & +.014 \\
Llama 3.2 3B & -.023 & -.176 & -.024 & .000 & -.003 & +.022 \\
Llama 3.1 8B & -.020 & -.202 & -.049 & .000 & -.003 & +.027 \\
Qwen3 1.7B & -.105 & -.620 & -.048 & .000 & -.002 & +.004 \\
Qwen3 4B & -.095 & -.491 & -.032 & .000 & -.002 & +.006 \\
Qwen3 8B & -.004 & -.691 & +.023 & .000 & .000 & +.001 \\
Gemma 3 1B & -.016 & -.222 & -.012 & -.003 & -.011 & +.029 \\
Gemma 3 4B & +.020 & -.204 & +.001 & -.003 & -.009 & +.014 \\
Gemma 3 12B & +.054 & -.180 & +.035 & -.004 & -.009 & +.005 \\
\bottomrule
\end{tabular}
\caption{Final-layer geometry change from base models to CPT and regularised CPT (+CR: cosine regularisation, +IS: I-STAR), averaged over 10 CPT languages.} %. $\Delta = \text{metric}_{\text{after}} - \text{metric}_{\text{before}}$
%\vspace{-0.3cm}
\label{tab:geometry_change}
\end{table}

%% file: tables/performance_summary.tex
\begin{table*}[t!]
\centering
\small
\begin{tabular}{ll rr rr rr r r r}
\toprule
Model & Variant & \multicolumn{2}{c}{AfriXNLI} & \multicolumn{2}{c}{AfriMMLU} & \multicolumn{2}{c}{AfriMGSM} & \multicolumn{3}{c}{Overall} \\
 &  & 0-shot & 5-shot & 0-shot & 5-shot & 0-shot & 8-shot & Avg & $\Delta$ & $\Delta$\% \\
\cmidrule(lr){3-4} \cmidrule(lr){5-6} \cmidrule(lr){7-8} \cmidrule(lr){9-11}
\midrule
Llama 3.2 1B & Base & 32.9 & 32.4 & \textbf{27.1} & 25.8 & 2.1 & \textbf{2.7} & 20.5 & -- & -- \\
 & CPT & \textbf{34.0} & \textbf{33.5} & 26.1 & \textbf{26.6} & 2.2 & 2.5 & \textbf{20.8} & \textbf{+0.3} & \textbf{+1.5\%} \\
 & + CosReg & 33.4 & 33.2 & 26.5 & \textbf{26.6} & \textbf{2.3} & 2.6 & \textbf{20.8} & \textbf{+0.3} & \textbf{+1.5\%} \\
 & + I-STAR & 33.9 & 33.2 & 26.0 & 26.5 & 2.2 & 2.6 & 20.7 & +0.2 & +1.0\% \\
\midrule
Gemma 3 1B & Base & 33.4 & 32.7 & \textbf{24.9} & 25.8 & 1.7 & 2.7 & 20.2 & -- & -- \\
 & CPT & 34.5 & \textbf{33.4} & 24.2 & \textbf{25.9} & \textbf{2.2} & 3.7 & \textbf{20.7} & \textbf{+0.5} & \textbf{+2.5\%} \\
 & + CosReg & 34.4 & 33.0 & 24.4 & 25.2 & 1.9 & \textbf{4.0} & 20.5 & +0.3 & +1.5\% \\
 & + I-STAR & \textbf{34.9} & 33.0 & 24.4 & 25.8 & 2.0 & 3.7 & 20.6 & +0.4 & +2.0\% \\
\midrule
Qwen3 1.7B & Base & 33.3 & 31.4 & \textbf{29.2} & 29.7 & \textbf{3.3} & \textbf{4.5} & \textbf{21.9} & -- & -- \\
 & CPT & 33.4 & \textbf{32.6} & 28.0 & 29.8 & 2.4 & 3.3 & 21.6 & -0.3 & -1.4\% \\
 & + CosReg & \textbf{33.9} & 32.3 & 28.2 & \textbf{30.0} & 2.6 & 3.2 & 21.7 & \textbf{-0.2} & \textbf{-0.9\%} \\
 & + I-STAR & 33.4 & \textbf{32.6} & 28.4 & 29.9 & 2.4 & 3.3 & 21.7 & \textbf{-0.2} & \textbf{-0.9\%} \\
\midrule
Llama 3.2 3B & Base & 33.6 & 32.3 & 26.4 & 27.5 & \textbf{2.7} & \textbf{4.1} & 21.1 & -- & -- \\
 & CPT & 33.8 & \textbf{32.7} & 26.5 & \textbf{28.4} & \textbf{2.7} & 4.0 & \textbf{21.3} & \textbf{+0.2} & \textbf{+0.9\%} \\
 & + CosReg & \textbf{33.9} & 32.4 & \textbf{27.0} & 28.0 & \textbf{2.7} & 3.9 & \textbf{21.3} & \textbf{+0.2} & \textbf{+0.9\%} \\
 & + I-STAR & \textbf{33.9} & 32.6 & 26.6 & 28.2 & 2.6 & 3.7 & \textbf{21.3} & \textbf{+0.2} & \textbf{+0.9\%} \\
\midrule
Qwen3 4B & Base & 33.6 & 32.1 & \textbf{31.8} & 34.4 & \textbf{4.7} & \textbf{6.6} & \textbf{23.9} & -- & -- \\
 & CPT & 33.8 & 33.0 & 30.9 & \textbf{34.8} & 3.1 & 6.5 & 23.7 & -0.2 & -0.8\% \\
 & + CosReg & 33.1 & \textbf{33.3} & 30.6 & 34.5 & 3.5 & 6.5 & 23.6 & -0.3 & -1.3\% \\
 & + I-STAR & \textbf{34.0} & \textbf{33.3} & 30.9 & 34.6 & 3.3 & 6.5 & 23.8 & \textbf{-0.1} & \textbf{-0.4\%} \\
\midrule
Gemma 3 4B & Base & 34.9 & 33.6 & 30.4 & 34.0 & 4.0 & 7.9 & 24.2 & -- & -- \\
 & CPT & 36.1 & 35.7 & 31.9 & 34.9 & 5.7 & 11.4 & 25.9 & +1.7 & +7.0\% \\
 & + CosReg & \textbf{36.9} & 35.6 & \textbf{32.7} & \textbf{35.1} & \textbf{6.1} & \textbf{11.5} & \textbf{26.3} & \textbf{+2.1} & \textbf{+8.7\%} \\
 & + I-STAR & 36.0 & \textbf{35.9} & 32.1 & 34.4 & 5.6 & 11.3 & 25.9 & +1.7 & +7.0\% \\
\midrule
Llama 3.1 8B & Base & 33.4 & 32.9 & 28.7 & 32.0 & 4.8 & 6.9 & 23.1 & -- & -- \\
 & CPT & 35.0 & 35.3 & \textbf{30.2} & \textbf{33.6} & 6.7 & \textbf{9.4} & 25.0 & +1.9 & +8.2\% \\
 & + CosReg & 35.3 & \textbf{36.1} & 29.9 & 33.4 & \textbf{6.8} & 9.2 & \textbf{25.1} & \textbf{+2.0} & \textbf{+8.7\%} \\
 & + I-STAR & \textbf{35.5} & 35.3 & 29.9 & 32.8 & 6.6 & 8.8 & 24.8 & +1.7 & +7.4\% \\
\midrule
Qwen3 8B & Base & \textbf{34.3} & 32.5 & 31.8 & 37.0 & \textbf{6.2} & 8.2 & 25.0 & -- & -- \\
 & CPT & 33.9 & 32.3 & \textbf{33.4} & 36.8 & 5.9 & 9.0 & 25.2 & +0.2 & +0.8\% \\
 & + CosReg & 33.6 & \textbf{32.6} & \textbf{33.4} & 37.0 & 5.9 & \textbf{9.6} & \textbf{25.3} & \textbf{+0.3} & \textbf{+1.2\%} \\
 & + I-STAR & 33.9 & 32.5 & 33.1 & \textbf{37.3} & 5.4 & 9.4 & \textbf{25.3} & \textbf{+0.3} & \textbf{+1.2\%} \\
\midrule
Gemma 3 12B & Base & 38.6 & 38.0 & 38.3 & 48.0 & 13.1 & 22.1 & 33.0 & -- & -- \\
 & CPT & 40.0 & \textbf{40.5} & \textbf{43.8} & \textbf{49.6} & 15.6 & \textbf{28.2} & 36.3 & +3.3 & +10.0\% \\
 & + CosReg & \textbf{40.1} & 40.3 & 43.7 & 49.5 & \textbf{16.5} & \textbf{28.2} & \textbf{36.4} & \textbf{+3.4} & \textbf{+10.3\%} \\
 & + I-STAR & \textbf{40.1} & \textbf{40.5} & 43.7 & 49.2 & 16.2 & 27.8 & 36.2 & +3.2 & +9.7\% \\
\bottomrule
\end{tabular}
\caption{Average performance (\%) across CPT languages (Base = pretrained model, CPT = vanilla continued pretraining, +CosReg / +I-STAR = regularised CPT). AfriMGSM scores are the averages of direct and chain-of-thought evaluations. We \textbf{boldface} the best-performing variant per model. $\Delta$ and $\Delta$\% show, respectively, absolute and relative performance gains over base models.}
\vspace{-0.3cm}
\label{tab:performance_summary}
\end{table*}

%% file: 5cpt.tex
%\section{Geometric Regularisation for Low-Resource Language Adaptation}

\section{CPT with Geometric Regularisation}
\label{sec:cpt_results}
In the previous section, we found that final-layer degeneration reliably worsens with data scarcity. % data scarcity reliably predicts final-layer representational degeneration
We therefore apply regularisation (Equations~\ref{eq:cos_reg} and~\ref{eq:istar_reg}) to final-layer token representations during CPT. We present our CPT experiments (Sections~\ref{subsec:cpt}--\ref{subsec:evaluation}), comparing the impact of vanilla CPT and geometrically regularised CPT on LLM geometry and downstream performance across 10 African languages.
% Seeing the final layer is where data scarcity has the greatest influence on LLM geometry, in the next section we focus on addressing degeneration in final-layer representations.

%\paragraph{Regularisation reduces final-layer degeneration.}

\subsection{CPT and Representational Degeneration}

First, we test whether geometric regularisation successfully changes LLM geometry as intended. Table~\ref{tab:geometry_change} confirms that it does. % -- we can explicitly steer the geometry of LLM representation spaces towards desired properties.
CosReg consistently decreases average pairwise cosine similarity in the final layer, so token representations become more locally separable. I-STAR consistently increases IsoScore in the final layer, so representation spaces utilise available dimensions more evenly. 
The effect of cosine similarity-based regularisation on isotropy, and vice versa, is negligible. This reaffirms that the two metrics measure different aspects of LLM geometry and that intervening on one does not affect the other. The effect of final-layer regularisation on other layers is also negligible (see Figures~\ref{fig:post_cpt_cosine} and~\ref{fig:post_cpt_isoscore} in the appendix), so our intervention is localised. We alter the geometry of a specific layer as intended, without disrupting the representational geometry of other layers.

Without regularisation, vanilla CPT already reduces final-layer degeneration, but only marginally. Table~\ref{tab:geometry_change} shows that final-layer cosine similarity decreases after vanilla CPT for most models, more so for the less multilingual models (Llama and Qwen) than for Gemma. Models learn more locally separable representations for low-resource languages, which might partially explain the success of CPT. Next, we study whether explicitly encouraging this geometric shift through regularisation can improve downstream performance over vanilla CPT.

\subsection{IrokoBench Results}

Table~\ref{tab:performance_summary} summarises the performance of our CPT models on IrokoBench, while Tables~\ref{tab:perf_xnli_mmlu} and~\ref{tab:perf_mgsm} in the appendix report the full set of results. Results are mixed across individual tasks and languages, but a clearer pattern emerges when results are viewed by model size, which is why Table~\ref{tab:performance_summary} orders models from smallest to largest. 
The final three columns of Table~\ref{tab:performance_summary} report average performance level and average gains over base models (in absolute and relative terms), across all tasks and languages.

\paragraph{CPT and geometric regularisation performance scales with model size.}
For smaller models (3B and below), geometrically regularised CPT offers no gains beyond vanilla CPT. However, at this scale, all forms of CPT achieve only marginal (or no) improvements over base models. With smaller models and limited pretraining data, CPT is generally less effective for language adaptation, and geometric regularisation fails to overcome this.

For larger models (4B and above), CPT more reliably improves over base models, and CosReg marginally improves average performance over vanilla CPT. While average gains are small, CosReg achieves notable improvements over vanilla CPT in several instances: AfriXNLI 5-shot for Llama 3.1 8B ($35.3 \rightarrow 36.1$), and AfriXNLI and AfriMMLU 0-shot for Gemma 3 4B ($36.1 \rightarrow 36.9$ and $31.9 \rightarrow 32.7$). On AfriMGSM, the most challenging task in IrokoBench, CosReg outperforms vanilla CPT across nearly all larger models, with notable gains in some cases ($9.0 \rightarrow 9.6$ for Qwen3 8B and $15.6 \rightarrow 16.5$ for Gemma 3 12B). At these low absolute performance levels, such improvements are proportionally substantial and reflect the average gain across 10 languages.
%At these low absolute performance levels, this represents proportionally substantial improvements based on an average across 10 languages.

% \paragraph{CosReg outperforms I-STAR.}
% Across models, average CosReg performance is higher than average I-STAR performance. 
% %I-STAR may benefit from different configurations -- for example, \citet{rudman2024stable} found that a negative $\lambda$ in Equation~\ref{eq:istar_reg} (encouraging anisotropy) improves finetuning performance. We experimented with this during hyperparameter tuning but did not observe improvements for CPT. 
% %We also cannot rule out that I-STAR might prove more effective with larger models, larger CPT corpora, or training from scratch.
% While we cannot rule out that I-STAR could benefit CPT in other configurations, such as larger target-language corpora if applied across all model layers , ...
% As shown in Table~\ref{tab:geometry_change}, both regularisers produce meaningful changes in final-layer geometry -- CosReg reduces cosine similarity and I-STAR increases isotropy as intended. 
% Therefore our results show that reducing cosine similarity in the final layer improves performance, while increasing isotropy does not. This suggests that, for low-resource language adaptation, local separability between token representations matters more than global utilisation of the embedding space.

\paragraph{CosReg outperforms I-STAR.}
Across models and tasks, CosReg tends to perform better than I-STAR. We cannot rule out that I-STAR could prove effective for CPT under different conditions, such as larger target-language corpora or if applied across multiple layers (the latter would require reducing the computational overhead of I-STAR). % which currently makes multi-layer regularisation prohibitively expensive during CPT)). 
However, the comparison between CosReg and I-STAR is informative because, as shown in Table~\ref{tab:geometry_change}, both regularisers alter model geometry as intended. 
Our results show that decreasing final-layer cosine similarity is a more reliable way to improve downstream performance than increasing final-layer isotropy. 
This suggests that, for low-resource language adaptation, local separability between token representations is more influential than global utilisation of the embedding space.

%% file: 6conclusion.tex
\section{Conclusion}

This work conducts a systematic analysis of LLM geometry, highlighting clear differences in how models construct representation spaces for low- and high-resource languages. 
Our results reveal a direct link between data scarcity and LLM geometry: lower-resource languages exhibit progressively worse representational degeneration, especially in final layers. 
Based on these findings, we adopt geometric regularisation techniques and apply them during CPT, where they successfully reduce final-layer degeneration for target languages. Cosine similarity-based regularisation frequently improves downstream performance for larger models, offering a simple technique to enhance CPT. %Our cosine similarity-based regulariser improves downstream performance over vanilla CPT, offering a simple technique to improve performance for low-resource languages.

Our approach is based on actionable interpretability \citep{mosbach-etal-2024-insights, orgad2026actionable}, which aims to use insights from interpretability experiments to inform modelling decisions. 
Our work shows how the perspective of LLM geometry is suited to this paradigm: geometric analyses can reveal internal deficiencies, while targeted geometric intervention can shift representational geometry towards desirable properties. More broadly, our work shows that actionable interpretability offers a promising approach to low-resource NLP. 
%Interpretability experiments can provide explanations for why models under-perform in low-resource settings, which can inform attempts to build data-efficient models. This allows researchers to move beyond trial-and-error towards more principled modelling decisions for low-resource languages.
By first understanding \emph{why} models underperform in low-resource settings, we can move beyond trial-and-error towards principled modelling decisions for low-resource languages.

\label{sec:conclusion}

%% file: 7limitations.tex
\section{Limitations}
\label{sec:limitations}

\paragraph{Analysis scope}
Our geometric analysis focusses on the relationship between language resource level and representational degeneration. 
While our results highlight the impact of other factors on LLM geometry, such as Amharic typography (Section~\ref{subsec:geometry_across_languages_and_layers}) and model architecture (Section~\ref{subsec:geometry_across_models}), we do not investigate these effects in detail. The role of linguistic typology and architectural choices in shaping LLM geometry warrants further study, as such insights could similarly inform strategies for improving low-resource language representations. We do not claim that data availability is the primary determinant of representational geometry, only that it is a strong and consistent one. Our main finding -- that cross-lingual geometric disparities scale with data availability -- holds across all 9 models and is robust when restricted to the Latin-script languages in our set, where script effects are controlled for.

\paragraph{Training scope}
The main practical limitation is that we focus on a single phase of LLM training, namely CPT. We cannot guarantee that geometric regularisation will effectively steer LLM geometry in other phases, such as training from scratch or instruction tuning. We initially applied geometric regularisation during instruction tuning of base models, but all forms of instruction tuning failed to reliably improve IrokoBench performance. Given the lack of high-quality instruction following data for African languages, this remains challenging for our target languages. We therefore focus on CPT, a reliable method for improving low-resource language performance and a natural setting for evaluating geometric intervention.

\paragraph{Performance gains}
The impact of geometric regularisation on performance is inconsistent across tasks, and most gains are small in absolute terms. 
CosReg marginally improves average performance over vanilla CPT, but only for larger models. Our primary practical contribution is not a new method for state-of-the-art CPT. Rather, we demonstrate that representational geometry is a measurable and actionable factor in low-resource language modelling. 
Our approach requires no additional data. Yet it reliably shifts final-layer geometry as intended, without degrading performance, and in some cases leads to meaningful gains on challenging tasks like AfriMGSM. 
This is sufficient to establish geometric regularisation as a promising direction for low-resource language adaptation.

\section{Acknowledgements}

This research was supported by Google.org and the Google Cloud Research Credits program for the Gemma Academic Program. Computations were also performed using facilities provided by the University of Cape Town’s ICTS High Performance Computing team: \url{hpc.uct.ac.za}.
This work is also based on research supported in part by the National Research Foundation of South Africa (Grant Number: 151601).

%% file: 8appendix.tex
\section{ Hyperparameters}
\label{appendix:hyperparams}

% We monolingually adapt base LLMs to 10 African languages on the WURA corpus \citep{oladipo-etal-2023-better}, with corpus sizes listed in Table~\ref{tab:wura_sizes}.

\subsection{Vanilla CPT}
\label{appendix:cpt_hyperparams}

We tune hyperparameters for vanilla CPT based on three pilot languages (Oromo, isiXhosa, isiZulu) using Llama 3.1 8B and Gemma 3 4B as representative models, selecting final settings based on AfriXNLI validation performance. We test two learning rates, $1\times10^{-5}$ and $1\times10^{-4}$. The higher rate performed worse, so we adopt $1\times10^{-5}$ with cosine decay to $1\times10^{-6}$. We train for 1, 2, and 3 epochs and find that performance plateaus after 1 epoch, so we train all subsequent models for 1 epoch only. We use a maximum sequence length of 512 tokens. These settings carry over unchanged to geometrically regularised CPT.

\subsection{Geometric regularisation}
\label{appendix:reg_hyperparams}

For both regularisers, we tune on the same three pilot languages, carrying over the CPT settings above. We select final hyperparameter values based on AfriXNLI validation performance and apply them uniformly to all other languages and models.

\paragraph{CosReg.} We tune the regularisation coefficient $\lambda \in \{-10, -1, 0.1, 1, 10\}$, where positive values penalise high cosine similarity (encouraging separation) and negative values do the opposite. We find $\lambda = 1$ performs best, which aligns with our expectation that penalising representational degeneration is beneficial. We also compare applying CosReg to the final layer only, to all layers, and to subsets of the final 3 and 5 layers, finding that regularising more layers fails to improve over final-layer only. 
%This is consistent with our analysis showing final-layer degeneration to be the most pronounced effect. 
We use $\lambda = 1$ applied to the final layer for all CosReg experiments.

\paragraph{I-STAR.} I-STAR introduces several additional hyperparameters, which we describe before detailing our tuning procedure.

\citet{rudman2024stable} propose I-STAR, a regularisation term based on IsoScore$^\star$, which is a differentiable variant of IsoScore \citep{rudman-etal-2022-isoscore}. 
%I-STAR computes an estimate of IsoScore from batch representations. 
Estimating covariance from one batch underestimates the true IsoScore of a representation space, which I-STAR addresses via shrinkage: interpolating the batch covariance with a stable reference covariance. At regular intervals, we compute a reference covariance matrix $C_0$ from a sample of $N$ token embeddings at layer $l$ drawn from WURA training data. At each training step, we compute the empirical covariance $C$ over a randomly sampled batch subset of $M'=1{,}000$ token embeddings at layer $l$, and form the shrinkage estimate:
\begin{align}
    \Sigma = (1-\zeta)C + \zeta C_0,
\end{align}
where $\zeta \in [0,1]$ controls the balance between the batch and the global reference estimate. IsoScore$^\star$ is computed from $\Sigma$, giving the training objective:
\begin{align}
    \mathcal{L}_{\text{I-STAR}} = \mathcal{L}_{\text{NLL}} + \lambda \cdot (1 - \text{IsoScore}^\star).
\end{align}
The sign of $\lambda$ determines the direction of the intervention: positive values encourage greater isotropy, while negative values encourage greater anisotropy. As with CosReg, the regularisation can be applied to a single layer or to all layers simultaneously.

\begin{table}[t!]
\centering
\small
\begin{tabular}{llrr}
\toprule
Code & Language & Words & Characters \\
\midrule
hau & Hausa         & 151,868,683 & 858,393,671 \\
amh & Amharic       &  86,144,742 & 453,397,345 \\
sot & Sesotho       &  35,331,080 & 193,620,015 \\
zul & isiZulu       &  31,584,011 & 271,658,420 \\
yor & Yor\`ub\'a    &  30,854,371 & 155,028,472 \\
ibo & Igbo          &  29,890,153 & 161,814,604 \\
sna & chiShona      &  28,464,659 & 216,818,585 \\
kin & Kinyarwanda   &  26,327,273 & 194,417,463 \\
xho & isiXhosa      &  13,082,332 & 111,534,373 \\
orm & Oromo         &   6,296,883 &  49,322,425 \\
\bottomrule
\end{tabular}
\caption{Corpus sizes for CPT (word counts are based on white-space separated tokens in the raw text).}
\label{tab:wura_sizes}
\end{table}

In all our experiments, we apply I-STAR regularisation to final-layer representations. 
Applying I-STAR to all layers requires computing IsoScore$^\star$ at every layer at each training step, which is prohibitively slow for the scale of our experiments. Selecting the final layer is also motivated by our analysis showing that the strongest representational degeneration effect, in terms of IsoScore, is found in the final layer (Table~\ref{tab:spearman_combined}).

\begin{figure*}[t!]
% \vspace{-0.5cm}
  \includegraphics[trim={0cm 0 0 0cm},clip,width=\linewidth]{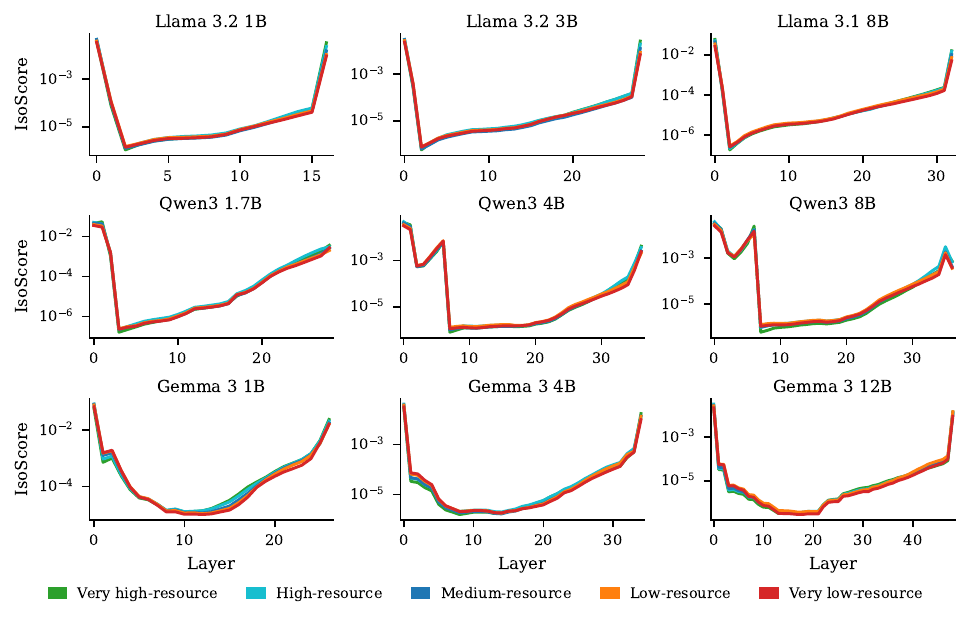}
  \caption{Isotropy (IsoScore) of representations across model layers, calculated as specified in Section~\ref{subsec:isotropy} and averaged across languages within each tier of data availability.} 
  %\vspace{-0.375cm}
  \label{fig:isoscore_layers}
\end{figure*}

%We tune three I-STAR hyperparameters. 
We tune the regularisation coefficient $\lambda \in \{-10, -1, 1, 10\}$, testing both signs to assess whether encouraging or discouraging isotropy is more beneficial. 
We also tune the shrinkage parameter $\zeta \in \{0.1, 0.2, 0.5, 0.8\}$, which controls how strongly the batch covariance is regularised towards the global reference. 
Lastly, we tune the reference sample size $N \in \{10{,}000, 50{,}000, 100{,}000, 250{,}000\}$, which determines how many token embeddings are used to compute the stable reference covariance $C_0$.

We find that $\lambda = 1$, $\zeta = 0.2$, and $N = 250{,}000$ perform best and use these values across all models and languages.

\begin{figure*}[h]
% \vspace{-0.5cm}
  \includegraphics[trim={0cm 0 0 0cm},clip,width=\linewidth]{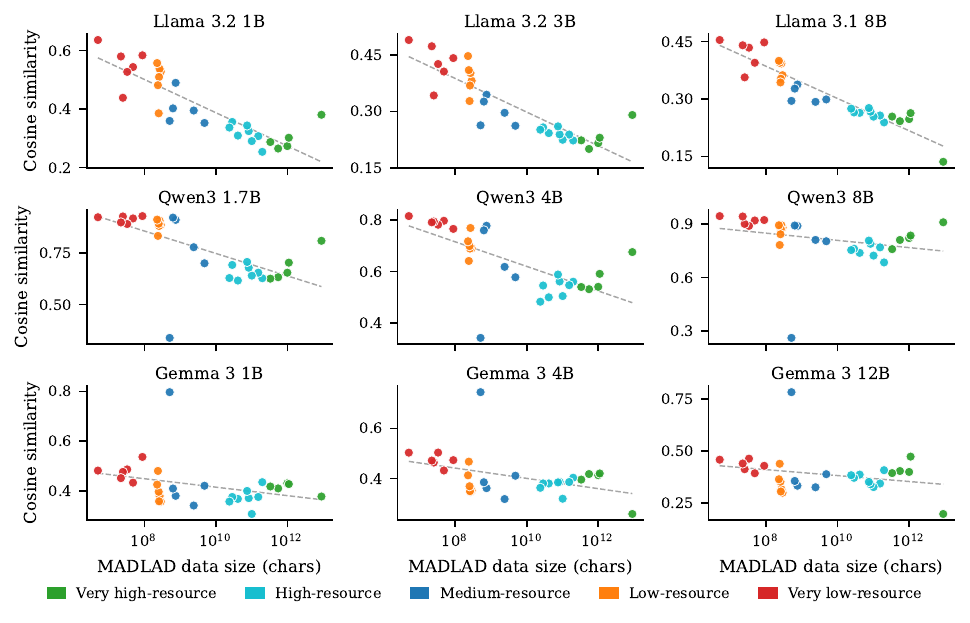}
  \caption{The relationship between pretraining data availability and average pairwise cosine similarity of final-layer representations. Each dot represents a single language.} 
  %\vspace{-0.375cm}
  \label{fig:full_final_layer_cosine}
    \vspace{0.5cm}
% \vspace{-0.5cm}
  \includegraphics[trim={0cm 0 0 0cm},clip,width=\linewidth]{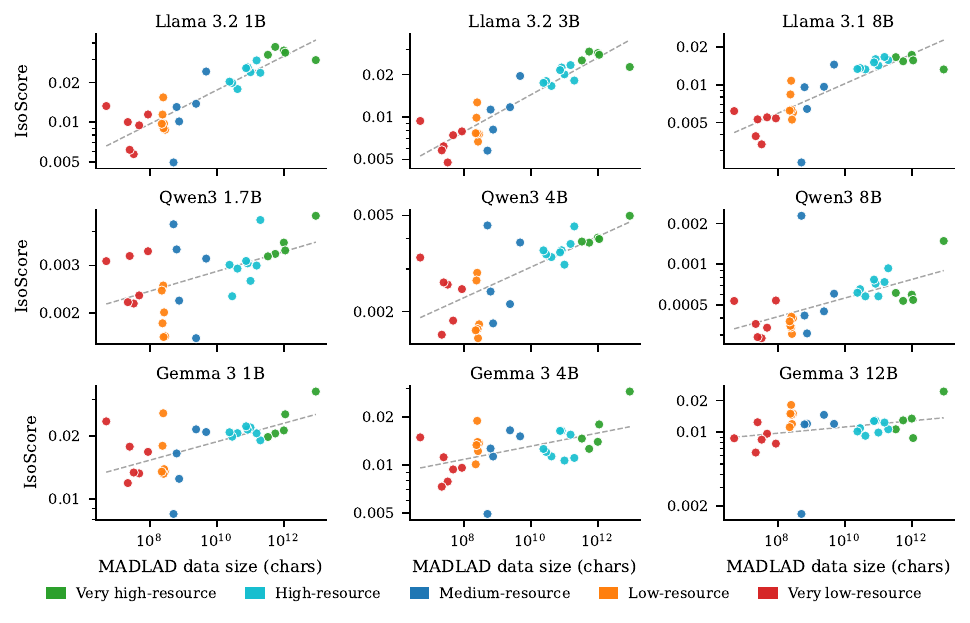}
  \caption{The relationship between pretraining data availability and isotropy (IsoScore) of final-layer representations. Each dot represents a single language.} 
  %\vspace{-0.375cm}
  \label{fig:full_final_layer_isoscore}
\end{figure*}

\begin{figure*}[t]
% \vspace{-0.5cm}
  \includegraphics[trim={0cm 0 0 0cm},clip,width=\linewidth]{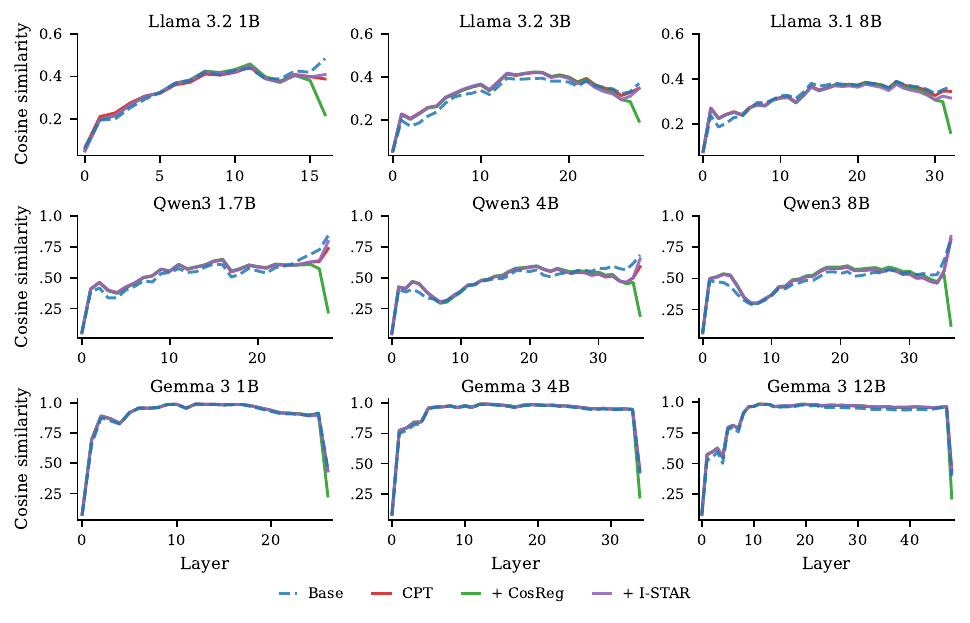}
  \vspace{-0.75cm}
  \caption{How do CPT and geometric regularisation change cosine similarity? We plot layer-wise cosine similarity, averaged across all 10 languages from our CPT experiments, comparing base models to vanilla CPT models (which only marginally affect cosine similarity), and geometrically regularised CPT models (CosReg successfully reduces final-layer cosine similarity, while leaving other layers unaffected).} 
  %\vspace{-0.375cm}
  \label{fig:post_cpt_cosine}
\end{figure*}

\begin{figure*}[t]
% \vspace{-0.5cm}
  \includegraphics[trim={0cm 0 0 0cm},clip,width=\linewidth]{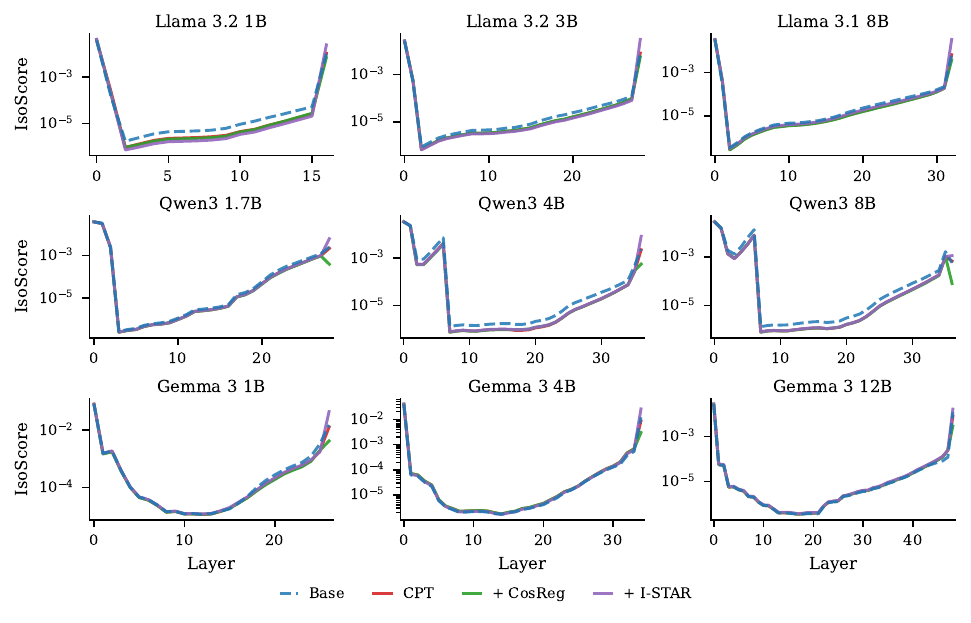}
  \vspace{-0.75cm}
  \caption{How do CPT and geometric regularisation change isotropy? We plot layer-wise IsoScore, averaged across all 10 languages from our CPT experiments, comparing base models to vanilla CPT models (which only marginally affect isotropy), and geometrically regularised CPT models (I-STAR successfully increases final-layer IsoScore, while leaving other layers unaffected).} 
  %\vspace{-0.375cm}
  \label{fig:post_cpt_isoscore}
\end{figure*}

\input{tables/performance_xnli_mmlu}
\input{tables/performance_mgsm}

%% file: tables/performance_xnli_mmlu.tex
\begin{table*}[t]
\centering
\tiny
\setlength{\tabcolsep}{3pt}
\begin{tabular}{ll|rrrrrrrrrrr|rrrrrrrrrrr}
\toprule
Model & Variant & kin & hau & amh & xho & sna & zul & ibo & yor & sot & orm & Avg & kin & hau & amh & xho & sna & zul & ibo & yor & sot & orm & Avg \\
\midrule
& & \multicolumn{11}{c|}{\textit{AfriXNLI} 0-shot (\%)} & \multicolumn{11}{c}{\textit{AfriXNLI} 5-shot (\%)} \\
\midrule
Llama 3.2 1B & Base & 33.3 & 34.5 & 32.8 & 32.3 & 31.3 & 33.0 & 32.7 & \textbf{34.2} & 33.7 & 30.8 & 32.9 & 31.7 & 33.7 & 30.7 & 31.7 & 29.2 & 32.3 & 32.7 & \textbf{34.7} & 32.8 & 32.5 & 32.4 \\
 & CPT & 33.3 & \textbf{34.8} & \textbf{33.0} & \textbf{33.5} & 33.7 & \textbf{36.2} & \textbf{33.8} & 33.8 & 33.3 & \textbf{33.5} & \textbf{34.0} & \textbf{34.3} & 33.7 & 33.8 & \textbf{34.2} & 31.3 & 35.5 & 33.2 & 32.8 & \textbf{33.8} & \textbf{33.0} & \textbf{33.5} \\
 & + CosReg & \textbf{33.7} & 34.5 & 32.7 & 33.3 & \textbf{34.5} & 33.0 & 33.5 & 33.0 & 31.5 & 33.3 & 33.4 & 33.2 & \textbf{35.0} & \textbf{34.2} & 33.7 & 31.2 & \textbf{35.7} & 32.5 & 32.2 & 32.7 & 32.5 & 33.2 \\
 & + I-STAR & 33.3 & 34.7 & 32.5 & 33.2 & 33.5 & 34.2 & 33.7 & 33.5 & \textbf{35.2} & \textbf{33.5} & 33.9 & 33.0 & 34.7 & 33.5 & 33.2 & \textbf{31.5} & 34.7 & \textbf{33.8} & 33.3 & 32.7 & 31.7 & 33.2 \\
\midrule
Llama 3.2 3B & Base & 33.2 & 34.2 & 33.2 & 33.3 & 33.3 & 34.5 & 33.8 & \textbf{33.0} & 33.5 & \textbf{34.0} & 33.6 & \textbf{33.3} & \textbf{33.7} & 34.3 & 31.8 & 30.8 & 33.7 & 31.8 & 29.2 & \textbf{33.0} & 33.2 & 32.3 \\
 & CPT & \textbf{33.3} & 34.2 & 35.0 & \textbf{33.7} & \textbf{34.3} & 34.8 & \textbf{34.2} & 32.5 & 33.3 & 33.5 & 33.8 & 31.5 & 33.0 & 34.5 & 32.7 & 33.5 & 34.7 & 31.7 & 30.5 & 32.8 & \textbf{34.3} & \textbf{32.7} \\
 & + CosReg & \textbf{33.3} & \textbf{35.3} & \textbf{35.2} & 33.5 & 34.2 & \textbf{35.7} & \textbf{34.2} & 32.0 & 33.3 & 33.3 & \textbf{33.9} & 30.2 & 33.0 & 33.3 & 32.7 & \textbf{34.3} & \textbf{35.8} & 31.8 & 29.7 & 30.2 & 33.8 & 32.4 \\
 & + I-STAR & \textbf{33.3} & 34.5 & 35.0 & \textbf{33.7} & 33.7 & 35.0 & \textbf{34.2} & 32.5 & \textbf{35.0} & 33.2 & \textbf{33.9} & 31.2 & 32.5 & \textbf{34.8} & \textbf{33.0} & 33.5 & 33.8 & \textbf{33.5} & \textbf{30.7} & 32.7 & 32.7 & 32.6 \\
\midrule
Llama 3.1 8B & Base & 32.5 & 35.2 & 34.3 & 33.7 & 32.3 & \textbf{35.5} & 33.5 & 30.5 & 34.2 & 33.3 & 33.4 & 32.3 & 32.7 & 31.2 & 33.7 & 33.3 & 33.2 & 32.8 & 29.8 & 32.8 & 35.7 & 32.9 \\
 & CPT & 32.7 & 40.0 & \textbf{34.5} & \textbf{35.0} & 36.2 & 32.7 & 34.0 & 37.0 & 35.0 & 32.8 & 35.0 & 31.5 & 39.0 & \textbf{37.3} & 34.8 & 35.3 & 37.2 & \textbf{33.5} & 34.7 & \textbf{36.5} & 35.2 & 35.3 \\
 & + CosReg & 34.2 & \textbf{40.5} & 34.2 & 33.2 & \textbf{37.3} & 34.3 & 34.0 & 35.8 & \textbf{35.3} & 33.3 & 35.3 & 32.0 & \textbf{40.0} & 36.3 & \textbf{36.8} & \textbf{35.7} & \textbf{38.0} & 32.8 & \textbf{37.3} & 35.7 & \textbf{36.7} & \textbf{36.1} \\
 & + I-STAR & \textbf{34.7} & 39.2 & 34.2 & 34.0 & 35.8 & 33.7 & \textbf{34.3} & \textbf{39.5} & 34.3 & \textbf{34.0} & \textbf{35.5} & \textbf{32.7} & 39.7 & 35.7 & 36.7 & 33.7 & 35.8 & 33.3 & 35.7 & 34.8 & 35.3 & 35.3 \\
\midrule
Qwen3 1.7B & Base & 33.5 & 33.3 & 35.2 & 33.2 & 32.8 & 33.2 & 33.2 & 32.5 & 33.7 & \textbf{34.0} & 33.3 & 28.7 & 32.3 & 32.3 & \textbf{33.2} & 30.2 & 31.8 & 31.8 & \textbf{33.0} & 29.3 & \textbf{32.5} & 31.4 \\
 & CPT & 33.3 & 32.3 & \textbf{36.3} & 33.3 & \textbf{34.7} & 32.7 & \textbf{33.5} & \textbf{35.2} & 32.8 & 33.0 & 33.4 & \textbf{32.2} & \textbf{33.2} & 34.3 & 32.7 & 34.7 & 32.5 & 31.8 & 32.3 & \textbf{32.5} & 31.2 & \textbf{32.6} \\
 & + CosReg & \textbf{33.8} & 33.5 & 35.2 & 33.3 & 34.0 & \textbf{34.8} & \textbf{33.5} & 33.5 & \textbf{35.3} & 33.2 & \textbf{33.9} & \textbf{32.2} & 32.5 & 34.0 & 33.0 & 34.5 & \textbf{34.3} & 31.5 & 31.3 & 30.3 & 30.8 & 32.3 \\
 & + I-STAR & 33.5 & \textbf{35.0} & 35.0 & \textbf{33.5} & 34.5 & 31.3 & 33.3 & 32.8 & 33.2 & 33.3 & 33.4 & 32.0 & 31.5 & \textbf{34.5} & 32.8 & \textbf{35.0} & 33.7 & \textbf{32.0} & 32.7 & \textbf{32.5} & 31.5 & \textbf{32.6} \\
\midrule
Qwen3 4B & Base & 32.7 & \textbf{33.3} & \textbf{37.5} & 33.2 & 34.0 & 33.7 & 33.7 & 34.5 & 33.5 & 33.5 & 33.6 & 31.2 & 30.3 & 34.3 & 32.7 & 32.3 & 32.8 & 31.2 & 33.2 & 31.5 & \textbf{34.0} & 32.1 \\
 & CPT & \textbf{33.5} & 32.0 & 34.8 & 33.5 & 34.3 & \textbf{34.7} & 33.5 & 34.5 & \textbf{34.5} & 33.3 & 33.8 & 33.5 & 31.3 & 35.7 & 34.0 & 31.8 & \textbf{35.0} & 30.7 & 34.5 & 33.7 & 32.7 & 33.0 \\
 & + CosReg & \textbf{33.5} & 32.5 & 34.2 & \textbf{34.0} & 31.3 & 33.2 & \textbf{33.8} & 32.7 & 33.5 & \textbf{33.8} & 33.1 & \textbf{34.2} & \textbf{33.0} & \textbf{38.5} & 34.2 & 31.8 & 32.7 & 32.5 & \textbf{34.7} & 33.5 & 33.2 & \textbf{33.3} \\
 & + I-STAR & \textbf{33.5} & 32.2 & 34.3 & 33.3 & \textbf{35.3} & 34.5 & 33.5 & \textbf{36.0} & 34.3 & 33.3 & \textbf{34.0} & 33.0 & 32.3 & 36.3 & \textbf{34.3} & \textbf{33.7} & 34.3 & \textbf{32.8} & 33.0 & \textbf{34.0} & 32.5 & \textbf{33.3} \\
\midrule
Qwen3 8B & Base & 32.2 & 34.8 & 33.2 & \textbf{33.8} & 32.5 & \textbf{35.8} & \textbf{34.2} & 34.5 & 34.0 & \textbf{36.5} & \textbf{34.3} & 32.3 & 32.3 & 37.3 & 31.8 & 31.7 & 34.3 & \textbf{31.7} & 31.3 & 33.3 & \textbf{33.5} & 32.5 \\
 & CPT & \textbf{33.2} & 34.2 & 33.7 & 33.5 & 33.3 & 33.8 & \textbf{34.2} & 32.3 & 34.2 & 36.3 & 33.9 & 32.2 & \textbf{32.5} & \textbf{39.0} & 33.8 & \textbf{32.5} & 33.8 & 29.8 & \textbf{33.2} & 31.7 & 31.0 & 32.3 \\
 & + CosReg & 30.3 & \textbf{38.2} & \textbf{34.2} & 32.3 & \textbf{33.8} & 34.0 & 31.8 & 32.7 & 34.3 & 35.2 & 33.6 & \textbf{34.0} & 32.2 & 38.3 & 33.3 & 31.7 & \textbf{35.0} & 31.5 & 30.5 & 33.7 & 31.5 & \textbf{32.6} \\
 & + I-STAR & 33.0 & 34.7 & 33.2 & 32.2 & 33.2 & 33.7 & 33.2 & \textbf{35.5} & \textbf{35.0} & 35.0 & 33.9 & 31.3 & 32.3 & 38.8 & \textbf{34.3} & 31.7 & 34.8 & 31.0 & 29.2 & \textbf{35.3} & 32.3 & 32.5 \\
\midrule
Gemma 3 1B & Base & 33.8 & 35.0 & \textbf{36.0} & 33.3 & 32.0 & 33.8 & 32.8 & 33.0 & 33.0 & 33.8 & 33.4 & 32.7 & 34.0 & 36.2 & 33.3 & 32.3 & 32.5 & \textbf{34.5} & 32.0 & 31.0 & 31.8 & 32.7 \\
 & CPT & \textbf{35.2} & 37.0 & 34.7 & 33.3 & 35.5 & 33.3 & \textbf{33.8} & 33.2 & 34.0 & 35.2 & 34.5 & \textbf{32.8} & 35.5 & 37.0 & \textbf{34.0} & 32.0 & 33.5 & 33.5 & \textbf{34.0} & \textbf{32.7} & \textbf{32.8} & \textbf{33.4} \\
 & + CosReg & 33.0 & 36.5 & 34.3 & \textbf{34.5} & 35.2 & 33.8 & \textbf{33.8} & \textbf{33.3} & 34.0 & \textbf{35.8} & 34.4 & 31.2 & \textbf{36.2} & \textbf{38.2} & \textbf{34.0} & 30.8 & 33.5 & 33.3 & 33.8 & 31.8 & 32.2 & 33.0 \\
 & + I-STAR & 33.7 & \textbf{38.2} & 35.0 & 34.0 & \textbf{36.2} & \textbf{34.2} & \textbf{33.8} & \textbf{33.3} & \textbf{36.0} & 34.8 & \textbf{34.9} & 32.2 & 35.8 & 36.3 & 32.8 & \textbf{33.0} & \textbf{33.8} & 32.7 & 33.2 & 31.3 & 32.3 & 33.0 \\
\midrule
Gemma 3 4B & Base & 35.2 & 38.0 & \textbf{38.7} & 34.3 & 36.5 & 33.2 & 31.8 & 35.8 & 35.8 & 33.8 & 34.9 & 32.7 & 36.0 & 39.8 & 35.8 & 34.8 & 33.7 & 31.2 & 33.7 & 34.2 & 30.8 & 33.6 \\
 & CPT & 34.8 & 39.2 & 37.0 & 37.3 & \textbf{42.5} & 33.3 & 31.3 & 34.8 & 37.0 & 34.5 & 36.1 & 35.2 & \textbf{41.5} & 40.5 & 36.8 & 34.7 & 35.5 & \textbf{35.7} & 33.5 & \textbf{35.8} & 32.7 & 35.7 \\
 & + CosReg & \textbf{37.5} & 39.8 & 36.5 & \textbf{37.8} & 41.0 & \textbf{34.3} & \textbf{35.0} & \textbf{36.7} & \textbf{37.2} & 33.0 & \textbf{36.9} & \textbf{35.8} & 41.3 & \textbf{41.3} & 35.8 & 35.2 & \textbf{35.8} & 34.5 & \textbf{35.2} & 34.5 & 32.5 & 35.6 \\
 & + I-STAR & 35.5 & \textbf{40.2} & 37.7 & 36.5 & 39.2 & 33.2 & 33.2 & 34.5 & 36.2 & \textbf{36.0} & 36.0 & 35.5 & 41.0 & 40.5 & \textbf{37.3} & \textbf{36.2} & 34.7 & 35.3 & 34.3 & 35.0 & \textbf{33.8} & \textbf{35.9} \\
\midrule
Gemma 3 12B & Base & 39.2 & \textbf{44.5} & 38.7 & 36.3 & 43.3 & \textbf{39.2} & 38.2 & 34.2 & 35.7 & \textbf{36.7} & 38.6 & 37.2 & 42.7 & 43.0 & 39.3 & 40.0 & \textbf{43.2} & 35.0 & 37.2 & 35.0 & 32.5 & 38.0 \\
 & CPT & 40.8 & 43.0 & 41.5 & 36.2 & 43.2 & 38.8 & \textbf{42.2} & 42.3 & 39.7 & 34.2 & 40.0 & 39.5 & 43.3 & \textbf{45.8} & 45.8 & \textbf{42.8} & 41.8 & 38.0 & \textbf{41.7} & \textbf{37.0} & 34.2 & \textbf{40.5} \\
 & + CosReg & 40.0 & 43.7 & 42.0 & \textbf{36.8} & 43.5 & 37.3 & 40.2 & \textbf{45.0} & \textbf{40.7} & 34.0 & \textbf{40.1} & 38.3 & 44.3 & 44.7 & \textbf{46.3} & 41.8 & 42.3 & 36.8 & 41.3 & 36.3 & \textbf{34.8} & 40.3 \\
 & + I-STAR & \textbf{41.2} & 43.3 & \textbf{42.2} & 36.0 & \textbf{44.7} & 38.5 & 41.0 & 42.2 & 39.0 & 34.8 & \textbf{40.1} & \textbf{40.0} & \textbf{45.3} & 44.0 & 45.2 & 42.5 & \textbf{43.2} & \textbf{38.2} & 39.7 & \textbf{37.0} & 33.2 & \textbf{40.5} \\
\midrule
& & \multicolumn{11}{c|}{\textit{AfriMMLU} 0-shot (\%)} & \multicolumn{11}{c}{\textit{AfriMMLU} 5-shot (\%)} \\
\midrule
Llama 3.2 1B & Base & 24.4 & \textbf{27.2} & \textbf{25.6} & 24.2 & \textbf{29.8} & 26.2 & \textbf{28.4} & 28.2 & \textbf{28.8} & \textbf{26.6} & \textbf{27.1} & 24.2 & 26.2 & 26.8 & 25.0 & 26.4 & 26.2 & 24.4 & 26.0 & 25.4 & \textbf{28.8} & 25.8 \\
 & CPT & \textbf{25.0} & 25.6 & 25.2 & 24.0 & 27.4 & 25.2 & 26.2 & \textbf{28.4} & 26.6 & 26.4 & 26.1 & \textbf{26.0} & 26.8 & 26.8 & \textbf{28.2} & 26.8 & 26.0 & \textbf{25.2} & \textbf{27.2} & 25.8 & 27.8 & \textbf{26.6} \\
 & + CosReg & 24.8 & 25.4 & \textbf{25.6} & \textbf{25.0} & 29.2 & \textbf{26.6} & 27.0 & 27.8 & 26.6 & 26.0 & 26.5 & 24.8 & 26.2 & \textbf{27.6} & 27.2 & \textbf{27.4} & 26.6 & 24.8 & 26.2 & \textbf{27.4} & \textbf{28.8} & \textbf{26.6} \\
 & + I-STAR & \textbf{25.0} & 25.0 & 25.2 & 24.0 & 28.2 & 25.8 & 25.6 & 27.8 & 26.4 & 26.0 & 26.0 & 25.2 & \textbf{27.4} & 25.4 & 27.0 & 26.6 & \textbf{27.0} & 24.2 & 26.8 & 25.2 & \textbf{28.8} & 26.5 \\
\midrule
Llama 3.2 3B & Base & 23.6 & 24.6 & 25.8 & 23.6 & 27.6 & 28.4 & 26.4 & \textbf{28.0} & 27.2 & 27.8 & 26.4 & 26.6 & 28.8 & 24.2 & 24.4 & 27.6 & \textbf{28.2} & 25.8 & \textbf{29.6} & 27.2 & 29.0 & 27.5 \\
 & CPT & 24.8 & 25.6 & \textbf{27.2} & 24.0 & 27.0 & 27.4 & \textbf{26.6} & 26.4 & 28.8 & 27.8 & 26.5 & \textbf{27.8} & 29.4 & \textbf{25.8} & \textbf{25.4} & \textbf{29.2} & 27.4 & \textbf{27.4} & 29.4 & \textbf{29.0} & 30.6 & \textbf{28.4} \\
 & + CosReg & 24.8 & \textbf{27.0} & 26.8 & \textbf{24.8} & 27.0 & \textbf{28.6} & 26.4 & 26.4 & \textbf{29.6} & 28.4 & \textbf{27.0} & 27.4 & \textbf{30.0} & \textbf{25.8} & 24.4 & 27.6 & 27.0 & 27.2 & 29.0 & 27.8 & \textbf{31.6} & 28.0 \\
 & + I-STAR & \textbf{25.0} & 26.4 & 26.2 & 24.6 & \textbf{27.8} & 26.4 & \textbf{26.6} & 26.2 & 27.8 & \textbf{28.8} & 26.6 & 27.2 & 29.6 & 25.4 & \textbf{25.4} & 28.8 & 26.6 & \textbf{27.4} & 28.6 & \textbf{29.0} & 31.0 & 28.2 \\
\midrule
Llama 3.1 8B & Base & 29.0 & 27.2 & 30.2 & 27.0 & 28.2 & 29.8 & 29.4 & 29.4 & 28.8 & 29.6 & 28.7 & 34.0 & 35.4 & 34.6 & 28.0 & 30.2 & 33.2 & 32.8 & 32.0 & 29.0 & 33.8 & 32.0 \\
 & CPT & 30.4 & 30.2 & 31.0 & 28.2 & 30.2 & \textbf{30.4} & \textbf{30.8} & \textbf{30.4} & 30.0 & \textbf{30.8} & \textbf{30.2} & 34.0 & \textbf{37.0} & 34.8 & 28.4 & \textbf{35.0} & \textbf{35.2} & \textbf{34.8} & \textbf{33.2} & 31.2 & 33.2 & \textbf{33.6} \\
 & + CosReg & 30.0 & \textbf{30.6} & \textbf{31.2} & 28.2 & 30.6 & 29.2 & 30.0 & 29.6 & \textbf{30.6} & 30.0 & 29.9 & \textbf{34.4} & 36.0 & \textbf{35.6} & \textbf{28.8} & 34.0 & 32.8 & 34.6 & 32.8 & \textbf{32.8} & \textbf{34.8} & 33.4 \\
 & + I-STAR & \textbf{31.2} & 30.0 & 29.4 & \textbf{28.6} & \textbf{30.8} & 30.0 & \textbf{30.8} & 29.2 & 29.4 & 28.8 & 29.9 & 33.6 & 36.0 & 33.0 & 27.6 & 33.8 & 33.6 & 33.6 & 31.8 & 30.8 & 34.0 & 32.8 \\
\midrule
Qwen3 1.7B & Base & \textbf{27.0} & 29.0 & 30.0 & 28.2 & \textbf{30.2} & \textbf{27.8} & \textbf{28.4} & \textbf{32.4} & \textbf{29.6} & \textbf{30.0} & \textbf{29.2} & \textbf{29.0} & \textbf{27.8} & 32.0 & \textbf{30.2} & 29.8 & 27.2 & 31.0 & 30.4 & 29.6 & \textbf{32.4} & 29.7 \\
 & CPT & 26.4 & 28.2 & 30.6 & 27.4 & 29.4 & 26.8 & 27.8 & 30.4 & 28.2 & 27.6 & 28.0 & 28.8 & 27.4 & 32.2 & 28.6 & \textbf{30.8} & 30.2 & 30.2 & 32.0 & 30.0 & 30.2 & 29.8 \\
 & + CosReg & 26.4 & \textbf{29.6} & 30.4 & 26.6 & 28.6 & 26.8 & 28.2 & 30.4 & 29.2 & 27.6 & 28.2 & 28.0 & \textbf{27.8} & \textbf{33.4} & 28.8 & 30.4 & 29.0 & \textbf{32.0} & 32.6 & \textbf{31.2} & 30.2 & \textbf{30.0} \\
 & + I-STAR & 26.4 & \textbf{29.6} & \textbf{30.8} & \textbf{28.4} & 29.2 & 27.2 & 27.0 & 30.8 & 28.4 & 28.4 & 28.4 & 28.6 & \textbf{27.8} & 32.2 & 29.6 & 29.8 & \textbf{31.2} & 30.2 & \textbf{33.0} & 29.6 & 29.6 & 29.9 \\
\midrule
Qwen3 4B & Base & \textbf{30.6} & \textbf{32.8} & \textbf{31.6} & \textbf{31.2} & 31.2 & 31.8 & \textbf{31.8} & 30.8 & \textbf{30.0} & \textbf{35.8} & \textbf{31.8} & \textbf{31.4} & 35.0 & 34.8 & \textbf{34.0} & 33.0 & 33.4 & 35.6 & 34.4 & 35.8 & \textbf{37.2} & 34.4 \\
 & CPT & 28.8 & 30.4 & 31.2 & 29.4 & \textbf{32.0} & \textbf{32.2} & 29.4 & \textbf{31.6} & \textbf{30.0} & 34.6 & 30.9 & 30.2 & \textbf{36.6} & 38.0 & 32.6 & 32.2 & 34.8 & \textbf{37.8} & 36.0 & \textbf{36.8} & 36.4 & \textbf{34.8} \\
 & + CosReg & 28.8 & 31.2 & \textbf{31.6} & 29.8 & 31.2 & 31.6 & 28.8 & 30.4 & 29.2 & 34.6 & 30.6 & 30.8 & 36.0 & \textbf{38.8} & 32.8 & 32.4 & \textbf{35.2} & \textbf{37.8} & 35.6 & 34.6 & 35.4 & 34.5 \\
 & + I-STAR & 28.8 & 30.2 & 30.4 & 29.2 & \textbf{32.0} & 31.6 & 30.0 & 31.2 & \textbf{30.0} & 35.0 & 30.9 & 29.8 & 36.0 & 38.2 & 32.8 & \textbf{33.2} & 34.4 & 36.6 & \textbf{36.4} & 36.0 & 36.4 & 34.6 \\
\midrule
Qwen3 8B & Base & 27.4 & 33.2 & 34.4 & 32.6 & 31.8 & 32.0 & 29.0 & 34.6 & 31.4 & 34.2 & 31.8 & \textbf{33.2} & 37.2 & 41.2 & \textbf{36.0} & \textbf{38.4} & 39.0 & 37.4 & \textbf{37.4} & 35.6 & 38.4 & 37.0 \\
 & CPT & 28.6 & \textbf{35.8} & 35.6 & \textbf{34.2} & 32.8 & 34.8 & 30.6 & 34.6 & 33.4 & \textbf{36.0} & \textbf{33.4} & \textbf{33.2} & 37.8 & 43.4 & 35.8 & 34.6 & 39.8 & 36.2 & 36.4 & 38.0 & 39.4 & 36.8 \\
 & + CosReg & \textbf{29.0} & 35.0 & \textbf{36.2} & 33.6 & 33.2 & \textbf{36.2} & \textbf{30.8} & 34.2 & \textbf{33.6} & 34.8 & \textbf{33.4} & 32.8 & \textbf{38.6} & 43.2 & 34.8 & 36.0 & \textbf{40.2} & 37.0 & 37.0 & 37.4 & 38.8 & 37.0 \\
 & + I-STAR & 28.0 & 34.8 & \textbf{36.2} & 33.4 & \textbf{33.4} & 35.2 & 30.2 & \textbf{35.0} & \textbf{33.6} & 34.2 & 33.1 & 32.4 & 37.0 & \textbf{43.8} & 35.6 & 35.8 & \textbf{40.2} & \textbf{38.4} & 37.0 & \textbf{39.4} & \textbf{40.0} & \textbf{37.3} \\
\midrule
Gemma 3 1B & Base & 22.4 & 22.8 & 21.2 & \textbf{26.0} & 23.6 & \textbf{25.2} & 24.2 & \textbf{27.6} & 27.6 & \textbf{25.0} & \textbf{24.9} & 24.6 & \textbf{24.6} & 22.8 & 27.8 & 24.6 & 26.2 & \textbf{25.4} & \textbf{25.6} & 27.6 & 26.2 & 25.8 \\
 & CPT & 22.6 & 23.0 & \textbf{22.6} & 24.4 & 24.8 & 23.6 & 24.8 & 24.8 & 27.2 & 22.6 & 24.2 & \textbf{25.6} & 23.8 & \textbf{23.2} & \textbf{28.0} & 25.2 & \textbf{27.6} & 23.8 & 24.6 & \textbf{28.6} & 26.0 & \textbf{25.9} \\
 & + CosReg & 22.6 & 23.0 & 20.8 & 25.6 & 24.6 & 23.8 & \textbf{25.2} & 24.8 & \textbf{27.8} & 21.8 & 24.4 & 24.2 & 22.8 & 22.4 & 25.6 & 26.0 & 26.0 & 24.8 & 24.2 & 27.4 & 26.2 & 25.2 \\
 & + I-STAR & \textbf{24.2} & \textbf{23.2} & \textbf{22.6} & 24.6 & \textbf{25.2} & 23.6 & 24.8 & 24.4 & 27.4 & 22.6 & 24.4 & 24.6 & 23.4 & 22.6 & 27.8 & \textbf{26.2} & 26.6 & 24.6 & 25.0 & 27.0 & \textbf{26.6} & 25.8 \\
\midrule
Gemma 3 4B & Base & 27.8 & 31.4 & 30.6 & 32.6 & 30.4 & 30.2 & 30.8 & 33.4 & 28.4 & 29.0 & 30.4 & 34.2 & 36.2 & 36.8 & 35.0 & 35.4 & \textbf{35.6} & 36.4 & 30.0 & 33.6 & 29.4 & 34.0 \\
 & CPT & \textbf{28.8} & \textbf{35.4} & 32.6 & 32.4 & 33.0 & 31.2 & \textbf{33.4} & 30.6 & 32.8 & 29.2 & 31.9 & \textbf{35.4} & \textbf{37.8} & \textbf{38.6} & 36.2 & 36.6 & 34.2 & 37.0 & 32.6 & 34.6 & 29.8 & 34.9 \\
 & + CosReg & 28.4 & 35.0 & 33.2 & \textbf{33.6} & \textbf{34.6} & 31.6 & 32.8 & \textbf{33.8} & \textbf{33.8} & \textbf{30.4} & \textbf{32.7} & 34.8 & \textbf{37.8} & 38.4 & \textbf{36.4} & \textbf{37.6} & 32.4 & 36.2 & \textbf{32.8} & \textbf{35.4} & \textbf{32.8} & \textbf{35.1} \\
 & + I-STAR & \textbf{28.8} & 35.0 & \textbf{34.2} & \textbf{33.6} & 32.8 & \textbf{32.2} & 33.0 & 31.2 & 32.4 & 30.0 & 32.1 & 35.2 & 37.2 & 37.8 & 35.4 & 36.4 & 32.0 & \textbf{37.4} & 32.2 & 34.2 & 29.8 & 34.4 \\
\midrule
Gemma 3 12B & Base & 39.8 & 42.2 & 44.2 & 37.8 & 36.2 & 39.2 & 42.2 & 35.6 & 38.0 & 33.6 & 38.3 & \textbf{47.4} & \textbf{52.8} & 53.4 & 46.6 & 52.2 & 49.2 & 48.2 & 44.8 & 50.0 & 41.0 & 48.0 \\
 & CPT & \textbf{42.4} & 48.8 & 48.6 & \textbf{46.0} & \textbf{44.6} & 41.4 & 45.4 & 42.2 & 44.8 & 38.6 & \textbf{43.8} & 46.8 & 51.2 & \textbf{56.4} & \textbf{50.2} & \textbf{55.4} & \textbf{50.8} & 48.6 & 45.0 & 53.2 & \textbf{44.8} & \textbf{49.6} \\
 & + CosReg & \textbf{42.4} & 48.4 & 48.2 & 45.2 & 44.0 & 40.8 & \textbf{46.4} & \textbf{43.8} & 44.4 & 37.8 & 43.7 & 45.4 & 52.0 & 55.2 & 49.2 & 53.8 & 50.2 & \textbf{49.8} & \textbf{46.4} & \textbf{53.8} & \textbf{44.8} & 49.5 \\
 & + I-STAR & 41.0 & \textbf{49.6} & \textbf{49.0} & 45.6 & 44.0 & \textbf{41.8} & 45.8 & 41.8 & \textbf{45.0} & \textbf{38.8} & 43.7 & 46.4 & 52.0 & 55.2 & 49.8 & 53.4 & \textbf{50.8} & 48.6 & 46.2 & 51.0 & \textbf{44.8} & 49.2 \\
\bottomrule
\end{tabular}
\caption{AfriXNLI and AfriMMLU performance (\%) across CPT languages. Best variants \textbf{boldfaced}.}
\label{tab:perf_xnli_mmlu}
\end{table*}

%% file: tables/performance_mgsm.tex
\begin{table*}[t]
\centering
\tiny
\setlength{\tabcolsep}{3pt}
\begin{tabular}{ll|rrrrrrrrrrr|rrrrrrrrrrr}
\toprule
Model & Variant & kin & hau & amh & xho & sna & zul & ibo & yor & sot & orm & Avg & kin & hau & amh & xho & sna & zul & ibo & yor & sot & orm & Avg \\
\midrule
& & \multicolumn{11}{c|}{\textit{AfriMGSM} 0-shot (\%)} & \multicolumn{11}{c}{\textit{AfriMGSM} 8-shot (\%)} \\
\midrule
Llama 3.2 1B & Base & 2.0 & 2.4 & 2.8 & 3.2 & \textbf{2.0} & 2.0 & \textbf{1.6} & \textbf{2.8} & 2.0 & \textbf{2.0} & 2.2 & \textbf{2.4} & 2.8 & \textbf{1.6} & 1.6 & 2.0 & 2.4 & \textbf{2.8} & \textbf{3.6} & 2.4 & \textbf{2.4} & 2.5 \\
 & CPT & \textbf{3.2} & \textbf{3.6} & 2.4 & 2.4 & \textbf{2.0} & 2.0 & 0.8 & 1.2 & 2.0 & \textbf{2.0} & 2.1 & 2.0 & \textbf{4.4} & 1.2 & \textbf{3.2} & \textbf{2.8} & 2.8 & 2.4 & 2.0 & 2.8 & 1.6 & 2.7 \\
 & + CosReg & 2.8 & 3.2 & 2.8 & \textbf{3.6} & 1.6 & \textbf{2.4} & 1.2 & 1.2 & 2.0 & \textbf{2.0} & 2.2 & \textbf{2.4} & 4.0 & 1.2 & \textbf{3.2} & \textbf{2.8} & \textbf{3.6} & 2.4 & 1.6 & \textbf{3.2} & 2.0 & \textbf{2.8} \\
 & + I-STAR & \textbf{3.2} & \textbf{3.6} & \textbf{3.2} & 2.8 & 1.6 & \textbf{2.4} & 0.8 & 1.6 & \textbf{2.4} & \textbf{2.0} & \textbf{2.3} & 2.0 & 4.0 & 1.2 & \textbf{3.2} & 2.4 & 2.8 & 2.4 & 1.6 & \textbf{3.2} & 2.0 & 2.6 \\
\midrule
Llama 3.2 3B & Base & 2.4 & \textbf{4.0} & \textbf{3.2} & \textbf{3.2} & 3.2 & \textbf{3.2} & \textbf{3.6} & \textbf{4.4} & \textbf{2.4} & 2.8 & \textbf{3.2} & \textbf{4.8} & \textbf{5.6} & 2.4 & \textbf{5.6} & 3.6 & 4.4 & 2.8 & \textbf{4.8} & \textbf{5.6} & 2.8 & \textbf{4.4} \\
 & CPT & \textbf{3.2} & \textbf{4.0} & 2.8 & 2.8 & \textbf{3.6} & 1.6 & 1.6 & 2.4 & \textbf{2.4} & 3.6 & 2.8 & \textbf{4.8} & 4.4 & \textbf{3.6} & 4.8 & 4.0 & 4.4 & 2.8 & 3.6 & 3.6 & 3.2 & 4.0 \\
 & + CosReg & \textbf{3.2} & 3.6 & 2.4 & 2.0 & 2.4 & 2.8 & 2.4 & 2.0 & \textbf{2.4} & \textbf{4.0} & 2.8 & \textbf{4.8} & 4.0 & 2.8 & 4.4 & \textbf{4.4} & \textbf{4.8} & \textbf{3.2} & 3.6 & 4.0 & 3.2 & 4.0 \\
 & + I-STAR & 2.4 & \textbf{4.0} & 2.8 & 2.4 & \textbf{3.6} & 2.4 & 2.0 & 2.4 & \textbf{2.4} & 3.6 & 2.8 & \textbf{4.8} & 4.0 & \textbf{3.6} & 4.0 & 4.0 & 4.4 & 2.8 & 3.6 & 4.0 & \textbf{3.6} & 3.9 \\
\midrule
Llama 3.1 8B & Base & 6.4 & 7.2 & 2.8 & 5.2 & 5.6 & 4.4 & 6.4 & 5.2 & 5.2 & 4.0 & 5.5 & \textbf{6.8} & 5.6 & 4.8 & \textbf{5.6} & \textbf{5.6} & 4.0 & \textbf{5.6} & \textbf{8.0} & \textbf{5.2} & \textbf{5.6} & \textbf{5.8} \\
 & CPT & \textbf{9.2} & \textbf{9.2} & 3.2 & 5.6 & \textbf{8.8} & \textbf{8.4} & 5.6 & 6.4 & 8.8 & \textbf{4.4} & 7.4 & 6.4 & \textbf{8.8} & 5.2 & 5.2 & 4.8 & \textbf{6.4} & 4.8 & 5.6 & \textbf{5.2} & 4.4 & 5.7 \\
 & + CosReg & \textbf{9.2} & 8.8 & 3.2 & 5.2 & 8.0 & \textbf{8.4} & 5.2 & \textbf{8.8} & \textbf{10.0} & \textbf{4.4} & \textbf{7.6} & 5.6 & \textbf{8.8} & 4.4 & \textbf{5.6} & 4.4 & 5.6 & 4.8 & 5.6 & 4.8 & 4.0 & 5.5 \\
 & + I-STAR & 8.8 & 7.6 & \textbf{4.8} & \textbf{6.0} & \textbf{8.8} & \textbf{8.4} & \textbf{7.2} & 8.0 & 8.4 & \textbf{4.4} & 7.5 & 5.6 & 7.6 & \textbf{6.0} & 5.2 & 5.2 & 6.0 & 4.0 & 5.2 & 4.8 & 4.4 & 5.3 \\
\midrule
Qwen3 1.7B & Base & \textbf{4.0} & \textbf{4.4} & \textbf{4.4} & \textbf{3.2} & \textbf{5.2} & 3.2 & \textbf{3.2} & \textbf{3.6} & \textbf{4.0} & \textbf{2.4} & \textbf{3.7} & \textbf{4.0} & \textbf{5.2} & 5.6 & \textbf{3.2} & \textbf{5.2} & \textbf{4.4} & \textbf{3.6} & \textbf{4.8} & \textbf{4.0} & \textbf{3.6} & \textbf{4.2} \\
 & CPT & 2.4 & 2.8 & \textbf{4.4} & 2.8 & 2.8 & 2.8 & 1.2 & 1.6 & 3.2 & \textbf{2.4} & 2.4 & 3.2 & 2.4 & 5.2 & \textbf{3.2} & 2.0 & 1.6 & 2.0 & 1.2 & 2.8 & \textbf{3.6} & 2.4 \\
 & + CosReg & 2.0 & 2.8 & 3.6 & 2.4 & 2.0 & 2.4 & 1.6 & 2.4 & 3.2 & \textbf{2.4} & 2.4 & 3.6 & 2.4 & 2.8 & 1.6 & 2.4 & 1.6 & 2.8 & 0.8 & 2.8 & 2.8 & 2.3 \\
 & + I-STAR & 2.4 & 2.8 & 4.0 & 2.8 & 2.4 & \textbf{4.0} & 1.2 & 1.6 & 3.2 & \textbf{2.4} & 2.5 & 3.2 & 2.8 & \textbf{6.0} & 2.4 & 2.4 & 2.0 & 2.0 & 1.2 & 2.8 & \textbf{3.6} & 2.5 \\
\midrule
Qwen3 4B & Base & \textbf{4.0} & \textbf{4.8} & \textbf{10.4} & \textbf{4.8} & \textbf{4.4} & \textbf{4.8} & \textbf{3.6} & \textbf{3.6} & \textbf{4.8} & \textbf{4.8} & \textbf{4.4} & 3.6 & 5.2 & 10.0 & \textbf{6.0} & \textbf{5.6} & 5.2 & \textbf{4.0} & \textbf{4.4} & \textbf{6.8} & \textbf{5.6} & \textbf{5.2} \\
 & CPT & 3.2 & 4.0 & 6.8 & 2.0 & 2.8 & 2.0 & 2.0 & 2.0 & 3.2 & 2.0 & 2.6 & \textbf{4.8} & \textbf{6.4} & \textbf{12.4} & 4.8 & 5.2 & \textbf{6.4} & 3.2 & \textbf{4.4} & 6.0 & 5.2 & \textbf{5.2} \\
 & + CosReg & 2.8 & 4.0 & 6.0 & 2.0 & 2.8 & 2.0 & 1.6 & 2.4 & 3.2 & 2.0 & 2.5 & 3.6 & 5.6 & 11.2 & 4.4 & \textbf{5.6} & 4.8 & 3.2 & 3.2 & 5.2 & 4.8 & 4.5 \\
 & + I-STAR & 2.8 & 2.4 & 6.8 & 2.4 & 2.8 & 2.4 & 1.6 & 2.0 & 4.0 & 2.0 & 2.5 & \textbf{4.8} & 5.6 & 12.0 & 4.8 & 5.2 & 5.2 & 3.2 & 3.6 & 6.4 & 4.8 & 4.8 \\
\midrule
Qwen3 8B & Base & \textbf{6.0} & 4.8 & \textbf{16.8} & \textbf{4.8} & \textbf{5.6} & \textbf{7.6} & \textbf{4.0} & \textbf{6.0} & \textbf{5.6} & 5.6 & \textbf{5.6} & \textbf{7.2} & 6.0 & \textbf{16.0} & 7.2 & \textbf{7.2} & 5.6 & \textbf{4.8} & \textbf{6.8} & 8.4 & \textbf{7.2} & \textbf{6.7} \\
 & CPT & 5.2 & \textbf{9.2} & 9.6 & \textbf{4.8} & 3.6 & 5.2 & 2.8 & 3.6 & 4.4 & \textbf{6.0} & 5.0 & 4.4 & 8.0 & 12.8 & 7.2 & 6.0 & 5.6 & 4.4 & 4.0 & 8.0 & 6.4 & 6.0 \\
 & + CosReg & 4.8 & 8.0 & 10.0 & \textbf{4.8} & 4.8 & 5.2 & 2.4 & 3.2 & 4.4 & 5.6 & 4.8 & 3.6 & \textbf{9.6} & 10.8 & 7.2 & 4.8 & \textbf{6.8} & 4.4 & 4.4 & 8.8 & 6.4 & 6.2 \\
 & + I-STAR & 5.6 & 7.2 & 8.8 & 4.4 & 4.0 & 4.0 & 2.8 & 3.6 & \textbf{5.6} & 4.8 & 4.7 & 4.0 & 8.4 & 13.2 & \textbf{8.0} & 5.6 & \textbf{6.8} & 4.0 & 4.8 & \textbf{9.2} & 6.0 & 6.3 \\
\midrule
Gemma 3 1B & Base & 1.6 & 2.0 & \textbf{2.0} & 1.2 & 2.8 & 1.2 & 1.2 & 0.8 & 1.2 & 1.6 & 1.5 & 2.4 & 3.6 & 3.6 & 2.8 & 3.2 & 3.2 & 2.4 & 3.6 & 3.6 & 3.2 & 3.1 \\
 & CPT & 1.6 & \textbf{2.8} & 1.6 & \textbf{2.0} & \textbf{3.6} & \textbf{2.0} & \textbf{1.6} & 1.2 & 1.6 & 2.0 & \textbf{2.0} & 5.2 & \textbf{5.2} & 6.0 & \textbf{4.8} & \textbf{4.8} & \textbf{4.8} & \textbf{5.2} & \textbf{5.6} & 4.4 & \textbf{6.0} & 5.1 \\
 & + CosReg & 1.6 & 1.2 & 1.2 & 1.2 & 1.6 & 1.6 & \textbf{1.6} & 1.2 & \textbf{2.8} & 2.0 & 1.6 & \textbf{6.0} & 4.8 & 4.8 & 4.4 & \textbf{4.8} & \textbf{4.8} & \textbf{5.2} & \textbf{5.6} & \textbf{4.8} & \textbf{6.0} & \textbf{5.2} \\
 & + I-STAR & \textbf{2.0} & 2.4 & 1.6 & 1.6 & 1.6 & 1.6 & \textbf{1.6} & \textbf{1.6} & 2.0 & \textbf{2.4} & 1.9 & 5.6 & \textbf{5.2} & \textbf{6.4} & 4.4 & \textbf{4.8} & \textbf{4.8} & 4.8 & \textbf{5.6} & 4.0 & \textbf{6.0} & 5.0 \\
\midrule
Gemma 3 4B & Base & 2.4 & 4.4 & \textbf{10.0} & 3.2 & 2.4 & 4.0 & 2.4 & 2.8 & 3.2 & 1.2 & 2.9 & 7.2 & 7.6 & 9.2 & 6.4 & 6.8 & 7.2 & 5.6 & 3.6 & 7.2 & 3.6 & 6.1 \\
 & CPT & 5.2 & 6.0 & 8.0 & 5.2 & 4.4 & 4.4 & \textbf{3.2} & 3.6 & \textbf{5.6} & 4.4 & 4.7 & 10.8 & 10.4 & \textbf{12.8} & 8.8 & 10.0 & 9.6 & \textbf{8.8} & \textbf{7.6} & 8.8 & 8.0 & 9.2 \\
 & + CosReg & \textbf{5.6} & \textbf{7.2} & 8.4 & \textbf{5.6} & \textbf{5.2} & \textbf{5.6} & \textbf{3.2} & \textbf{5.2} & 4.4 & \textbf{5.6} & \textbf{5.3} & 10.4 & 10.4 & 12.0 & \textbf{9.2} & \textbf{10.8} & \textbf{10.4} & \textbf{8.8} & \textbf{7.6} & \textbf{9.2} & \textbf{8.8} & \textbf{9.5} \\
 & + I-STAR & 4.0 & 6.8 & 7.2 & 4.8 & 4.8 & 4.8 & 2.4 & 4.0 & 4.0 & 4.4 & 4.4 & \textbf{11.2} & \textbf{10.8} & 12.4 & 8.0 & 10.0 & 9.6 & \textbf{8.8} & 6.8 & \textbf{9.2} & 8.4 & 9.2 \\
\midrule
Gemma 3 12B & Base & 14.0 & 15.6 & \textbf{19.2} & 11.2 & 11.2 & 12.4 & 8.8 & 7.6 & 11.6 & 8.0 & 11.2 & 20.8 & 20.8 & 21.6 & 18.4 & \textbf{23.2} & 20.0 & 14.4 & 12.8 & 19.6 & 14.4 & 18.3 \\
 & CPT & 15.6 & 16.0 & 14.8 & 13.2 & 15.2 & 17.6 & \textbf{11.2} & 12.4 & 16.8 & 15.2 & 14.8 & \textbf{24.4} & \textbf{23.2} & \textbf{24.4} & \textbf{20.4} & 22.4 & \textbf{24.4} & 17.6 & 17.2 & \textbf{23.6} & 20.8 & \textbf{21.6} \\
 & + CosReg & \textbf{17.6} & \textbf{16.4} & 13.6 & 12.8 & 16.4 & \textbf{18.0} & 10.4 & 13.2 & \textbf{17.6} & \textbf{16.8} & 15.5 & 23.2 & \textbf{23.2} & 22.0 & 17.6 & 22.4 & 21.2 & \textbf{18.0} & 17.2 & \textbf{23.6} & \textbf{22.4} & 21.0 \\
 & + I-STAR & 16.0 & \textbf{16.4} & 14.0 & \textbf{14.4} & \textbf{17.6} & \textbf{18.0} & 10.4 & \textbf{13.6} & \textbf{17.6} & \textbf{16.8} & \textbf{15.6} & 24.0 & 22.4 & 24.0 & 18.8 & 22.0 & 23.6 & 17.6 & \textbf{18.8} & 22.4 & \textbf{22.4} & 21.3 \\
\midrule
& & \multicolumn{11}{c|}{\textit{AfriMGSM CoT} 0-shot (\%)} & \multicolumn{11}{c}{\textit{AfriMGSM CoT} 8-shot (\%)} \\
\midrule
Llama 3.2 1B & Base & \textbf{2.8} & 1.2 & \textbf{2.0} & 2.0 & 2.0 & \textbf{2.4} & \textbf{2.0} & \textbf{2.4} & 1.6 & \textbf{2.0} & 2.0 & 3.2 & 2.0 & \textbf{2.0} & \textbf{3.2} & \textbf{4.8} & 2.0 & \textbf{2.0} & \textbf{3.2} & 3.2 & \textbf{2.4} & \textbf{2.9} \\
 & CPT & 2.0 & 1.6 & 1.6 & \textbf{3.2} & \textbf{2.8} & 2.0 & \textbf{2.0} & 2.0 & 2.0 & \textbf{2.0} & 2.2 & 2.8 & \textbf{2.8} & 1.6 & 2.8 & 2.8 & 2.4 & 1.2 & 2.4 & 2.4 & 1.6 & 2.4 \\
 & + CosReg & \textbf{2.8} & \textbf{2.0} & 1.6 & \textbf{3.2} & 2.0 & \textbf{2.4} & \textbf{2.0} & \textbf{2.4} & \textbf{2.4} & \textbf{2.0} & \textbf{2.4} & 2.4 & 2.4 & 1.6 & \textbf{3.2} & 2.8 & \textbf{2.8} & 1.2 & 1.2 & \textbf{3.6} & 2.0 & 2.4 \\
 & + I-STAR & 2.4 & \textbf{2.0} & \textbf{2.0} & 2.4 & 2.0 & 1.6 & 1.6 & \textbf{2.4} & 2.0 & \textbf{2.0} & 2.0 & \textbf{3.6} & \textbf{2.8} & 0.8 & \textbf{3.2} & 2.8 & 2.0 & 1.2 & 2.8 & 2.4 & 1.6 & 2.5 \\
\midrule
Llama 3.2 3B & Base & 2.0 & 2.0 & 2.0 & \textbf{3.2} & 2.0 & \textbf{3.2} & 1.2 & 1.6 & \textbf{2.8} & 2.0 & 2.2 & 2.8 & \textbf{6.0} & \textbf{3.6} & 3.2 & 3.6 & 2.4 & \textbf{2.4} & 3.6 & \textbf{4.8} & 4.4 & 3.7 \\
 & CPT & 2.8 & 2.8 & \textbf{2.8} & 2.4 & \textbf{2.8} & 1.6 & 2.4 & 2.4 & 2.4 & \textbf{3.2} & 2.5 & 4.0 & 3.2 & 2.4 & \textbf{5.2} & \textbf{4.8} & \textbf{4.4} & 1.6 & 4.0 & 3.6 & \textbf{5.2} & \textbf{4.0} \\
 & + CosReg & \textbf{3.2} & 2.4 & 2.0 & 2.8 & 2.4 & 2.0 & \textbf{3.2} & \textbf{2.8} & 2.0 & 2.4 & \textbf{2.6} & 3.6 & 2.8 & 2.8 & 3.2 & 4.0 & 4.0 & \textbf{2.4} & \textbf{4.8} & 4.0 & 4.8 & 3.7 \\
 & + I-STAR & 2.0 & \textbf{3.2} & 2.4 & 2.8 & \textbf{2.8} & 1.2 & 1.6 & 2.0 & 2.0 & \textbf{3.2} & 2.3 & \textbf{4.4} & 3.2 & 2.4 & 2.8 & 4.4 & 4.0 & 2.0 & 4.0 & 2.0 & 4.4 & 3.5 \\
\midrule
Llama 3.1 8B & Base & 4.4 & 7.6 & \textbf{3.6} & 3.6 & 3.6 & 2.0 & \textbf{4.4} & 3.6 & 4.8 & 2.0 & 4.0 & 7.6 & 15.6 & 3.6 & 4.8 & 8.0 & 6.8 & 8.4 & 6.8 & 7.2 & 6.4 & 8.0 \\
 & CPT & 5.6 & 9.2 & \textbf{3.6} & 4.4 & 8.8 & \textbf{7.6} & \textbf{4.4} & 4.4 & 6.4 & 4.0 & \textbf{6.1} & 14.0 & \textbf{16.8} & \textbf{5.6} & 10.8 & \textbf{15.2} & 12.0 & \textbf{14.0} & \textbf{11.2} & 14.4 & 8.8 & \textbf{13.0} \\
 & + CosReg & \textbf{6.4} & \textbf{10.0} & 2.8 & \textbf{4.8} & \textbf{9.6} & 4.8 & \textbf{4.4} & \textbf{4.8} & \textbf{7.6} & 2.4 & \textbf{6.1} & 14.0 & 14.0 & 4.8 & \textbf{12.4} & 14.4 & \textbf{12.8} & 13.2 & \textbf{11.2} & \textbf{15.2} & 8.8 & 12.9 \\
 & + I-STAR & \textbf{6.4} & 8.0 & 2.8 & \textbf{4.8} & 8.4 & 5.2 & 4.0 & \textbf{4.8} & 5.6 & \textbf{4.4} & 5.7 & \textbf{14.4} & 12.8 & 4.4 & 9.6 & 14.0 & 10.4 & 12.8 & \textbf{11.2} & 14.4 & \textbf{10.0} & 12.2 \\
\midrule
Qwen3 1.7B & Base & \textbf{3.6} & 2.8 & \textbf{7.2} & 3.2 & \textbf{3.6} & \textbf{2.8} & 3.6 & 2.4 & \textbf{2.0} & \textbf{2.4} & \textbf{2.9} & \textbf{4.0} & 5.2 & 3.6 & 3.2 & \textbf{6.4} & \textbf{6.4} & \textbf{3.2} & 5.2 & \textbf{4.8} & 4.4 & \textbf{4.8} \\
 & CPT & 2.0 & \textbf{3.2} & 4.4 & 3.2 & 3.2 & 2.4 & 2.4 & 2.0 & 1.2 & \textbf{2.4} & 2.4 & 3.2 & \textbf{5.6} & 4.0 & \textbf{4.4} & 4.4 & 3.2 & 2.4 & \textbf{6.4} & 2.8 & \textbf{5.2} & 4.2 \\
 & + CosReg & 2.4 & 2.8 & 3.6 & \textbf{4.0} & 2.0 & \textbf{2.8} & 3.6 & \textbf{3.2} & \textbf{2.0} & \textbf{2.4} & 2.8 & 3.2 & \textbf{5.6} & 2.4 & 3.2 & 5.2 & 4.8 & 2.4 & 5.6 & 3.6 & 3.6 & 4.1 \\
 & + I-STAR & 2.4 & 2.0 & 4.0 & 2.0 & 2.4 & 1.6 & \textbf{4.4} & 2.4 & 1.6 & 2.0 & 2.3 & \textbf{4.0} & 4.4 & \textbf{4.4} & \textbf{4.4} & 4.4 & 2.8 & 2.4 & 6.0 & 4.0 & 4.8 & 4.1 \\
\midrule
Qwen3 4B & Base & \textbf{3.6} & 7.6 & \textbf{7.6} & \textbf{5.2} & \textbf{6.0} & \textbf{4.8} & \textbf{3.2} & \textbf{4.4} & \textbf{6.4} & 4.4 & \textbf{5.1} & \textbf{8.4} & 8.4 & 10.8 & \textbf{7.6} & 8.8 & 8.4 & 2.8 & 7.2 & \textbf{10.4} & \textbf{10.8} & 8.1 \\
 & CPT & 2.0 & 9.2 & 4.8 & 3.2 & 3.6 & 2.0 & 2.4 & 2.8 & 3.2 & 4.0 & 3.6 & 6.0 & 12.4 & 10.4 & 6.8 & 8.4 & 8.4 & \textbf{3.2} & 7.2 & 9.2 & 9.2 & 7.9 \\
 & + CosReg & 3.2 & \textbf{10.4} & 3.6 & 3.2 & 4.0 & 4.4 & 2.8 & 4.0 & 2.8 & \textbf{5.6} & 4.5 & 7.6 & \textbf{15.2} & 8.8 & 7.2 & \textbf{9.6} & 8.8 & 2.8 & 7.6 & 8.8 & 8.4 & \textbf{8.4} \\
 & + I-STAR & 2.0 & 9.2 & 4.4 & 3.6 & 5.2 & 4.0 & 2.4 & 3.2 & 2.4 & \textbf{5.6} & 4.2 & \textbf{8.4} & 13.2 & \textbf{11.2} & 6.4 & 8.0 & \textbf{9.2} & 2.8 & \textbf{8.0} & 8.0 & 9.2 & 8.1 \\
\midrule
Qwen3 8B & Base & 7.6 & 6.4 & \textbf{19.2} & \textbf{7.6} & 6.8 & \textbf{8.4} & \textbf{4.0} & \textbf{5.2} & 6.8 & \textbf{8.8} & 6.8 & 10.4 & 8.8 & 15.2 & 10.8 & 9.2 & 9.2 & 3.6 & 9.2 & 12.4 & 13.2 & 9.6 \\
 & CPT & 6.8 & \textbf{12.4} & 8.4 & 3.2 & 7.2 & 8.0 & 3.6 & \textbf{5.2} & \textbf{9.6} & 5.2 & 6.8 & 10.8 & 16.8 & \textbf{16.8} & 10.8 & 10.8 & 12.0 & 5.6 & \textbf{10.8} & \textbf{15.6} & 14.0 & 11.9 \\
 & + CosReg & \textbf{8.4} & 10.0 & 9.2 & 3.6 & \textbf{9.2} & 7.2 & 3.2 & 4.8 & 9.2 & 6.8 & \textbf{6.9} & \textbf{11.2} & \textbf{23.2} & 15.6 & \textbf{12.4} & \textbf{13.2} & 12.0 & \textbf{6.0} & 10.4 & 13.6 & \textbf{14.8} & \textbf{13.0} \\
 & + I-STAR & 6.0 & 10.0 & 8.0 & 4.4 & 6.8 & 6.0 & 3.6 & 4.4 & 8.8 & 5.6 & 6.2 & 10.8 & \textbf{23.2} & \textbf{16.8} & 10.8 & 10.8 & \textbf{12.8} & 4.8 & \textbf{10.8} & 14.8 & 12.8 & 12.4 \\
\midrule
Gemma 3 1B & Base & 3.2 & 1.6 & \textbf{2.4} & 2.0 & 1.6 & 1.6 & \textbf{1.6} & 2.0 & 1.6 & 1.6 & 1.9 & 3.6 & \textbf{2.8} & \textbf{3.2} & 1.6 & \textbf{3.2} & 1.6 & 2.0 & 1.6 & 1.2 & \textbf{2.8} & 2.3 \\
 & CPT & \textbf{4.0} & \textbf{3.2} & 1.6 & 1.6 & 2.4 & 2.0 & 1.2 & 2.0 & \textbf{2.0} & 2.4 & \textbf{2.3} & 3.2 & 1.6 & 2.8 & 2.8 & 2.0 & 2.8 & 2.8 & \textbf{2.4} & \textbf{2.0} & 1.6 & 2.4 \\
 & + CosReg & 2.8 & 2.4 & 1.6 & 2.0 & \textbf{3.2} & 1.6 & 0.8 & 1.6 & \textbf{2.0} & \textbf{3.2} & 2.2 & \textbf{4.4} & 2.4 & 2.0 & \textbf{3.2} & \textbf{3.2} & \textbf{4.0} & 2.0 & 1.6 & 1.2 & \textbf{2.8} & \textbf{2.8} \\
 & + I-STAR & 3.2 & 2.0 & 1.2 & \textbf{2.4} & 1.6 & \textbf{2.8} & 1.2 & \textbf{2.4} & 1.6 & 2.4 & 2.2 & 2.4 & 2.0 & 2.8 & \textbf{3.2} & 2.8 & 2.0 & \textbf{3.2} & 2.0 & 1.6 & 2.0 & 2.4 \\
\midrule
Gemma 3 4B & Base & 5.2 & 7.6 & \textbf{10.8} & 6.8 & 4.4 & 5.2 & \textbf{5.6} & 2.8 & 5.6 & 3.2 & 5.2 & 9.2 & 15.2 & 19.2 & 10.0 & 14.4 & 12.8 & 5.6 & 6.0 & 10.4 & 4.0 & 9.7 \\
 & CPT & \textbf{10.0} & 9.6 & 6.4 & 8.0 & 7.6 & \textbf{7.6} & 3.6 & 4.0 & 6.0 & 3.6 & 6.7 & 15.2 & 20.0 & 18.4 & 13.6 & \textbf{18.4} & 15.6 & 10.4 & 5.6 & \textbf{14.8} & \textbf{8.8} & \textbf{13.6} \\
 & + CosReg & 9.2 & \textbf{10.0} & 6.8 & 6.8 & \textbf{9.2} & 6.8 & 2.8 & \textbf{5.2} & \textbf{7.2} & \textbf{5.6} & \textbf{7.0} & \textbf{16.4} & \textbf{21.2} & 20.4 & 13.2 & 18.0 & \textbf{16.4} & 8.8 & 6.0 & 13.2 & 8.4 & 13.5 \\
 & + I-STAR & 9.6 & 9.6 & 7.2 & \textbf{8.4} & 6.8 & 6.8 & 3.6 & 4.4 & 6.4 & 4.4 & 6.7 & 15.6 & 18.4 & \textbf{21.2} & \textbf{14.4} & 18.0 & 14.4 & \textbf{10.8} & \textbf{6.4} & 14.4 & 8.0 & 13.4 \\
\midrule
Gemma 3 12B & Base & 16.8 & \textbf{23.6} & 23.2 & 11.6 & 14.4 & 16.4 & 10.0 & 14.4 & \textbf{18.8} & 10.0 & 15.1 & 26.8 & 37.2 & \textbf{46.4} & 27.6 & 33.2 & 30.4 & 24.0 & 7.6 & 29.6 & 17.6 & 26.0 \\
 & CPT & 23.2 & 22.4 & 21.6 & 9.6 & 16.4 & 16.0 & \textbf{11.6} & \textbf{18.0} & 14.0 & 16.4 & 16.4 & \textbf{39.6} & 41.2 & 40.8 & 31.2 & \textbf{38.0} & 37.6 & \textbf{29.2} & 28.4 & 36.4 & 32.8 & 34.9 \\
 & + CosReg & \textbf{23.6} & 21.2 & \textbf{24.0} & \textbf{16.8} & \textbf{17.6} & 15.6 & 10.4 & \textbf{18.0} & 16.0 & \textbf{19.2} & \textbf{17.6} & \textbf{39.6} & \textbf{42.0} & 42.8 & 30.4 & 34.4 & 38.0 & 28.8 & \textbf{30.8} & \textbf{41.6} & 32.8 & \textbf{35.4} \\
 & + I-STAR & 19.2 & 23.2 & 22.0 & 10.4 & \textbf{17.6} & \textbf{19.6} & 10.8 & 17.6 & 13.6 & 18.0 & 16.7 & 38.4 & 40.4 & 42.4 & \textbf{31.6} & 36.4 & \textbf{38.4} & 26.4 & 26.8 & 37.2 & \textbf{33.2} & 34.3 \\
\bottomrule
\end{tabular}
\caption{AfriMGSM direct and chain-of-thought performance (\%) across CPT languages. Best variants \textbf{boldfaced}.}
\label{tab:perf_mgsm}
\end{table*}